\pdfoutput=1
\documentclass[journal]{IEEEtran}

\usepackage[ruled,vlined]{algorithm2e}

\usepackage{comment}
\usepackage{subfigure}
\usepackage{graphicx,epsfig}
\def\BibTeX{{\rm B\kern-.05em{\sc i\kern-.025em b}\kern-.08em
    T\kern-.1667em\lower.7ex\hbox{E}\kern-.125emX}}

\usepackage{mathtools, cuted}
\usepackage{breqn}
\usepackage{autobreak}
\usepackage{diagbox}
\usepackage[noadjust]{cite}

\newcommand{\calD}{\mathcal{D}}
\newcommand{\calS}{\mathcal{S}}

\newcommand{\calL}{\mathcal{L}}
\newcommand{\calH}{\mathcal{H}}

\newcommand{\calZ}{\mathcal{Z}}

\newcommand{\Z}{\mathbf{Z}}

\newcommand{\G}{\mathbf{G}}

\newcommand{\bpi}{\boldsymbol{\pi}}

\newcommand{\btheta}{\boldsymbol{\theta}}

\newcommand{\bTheta}{\boldsymbol{\Theta}}

\begin{document}

\title{Crowding Prediction of In-Situ Metro Passengers Using Smart Card Data}

\author{Xiancai Tian, Chen Zhang, Baihua Zheng
\thanks{This research is supported by the National Research Foundation, Singapore under its International Research Centres in Singapore Funding Initiative. Any opinions, findings and conclusions or recommendations expressed in this material are those of the author(s) and do not reflect the views of National Research Foundation, Singapore.}
\thanks{X. Tian is with Living Analytics Research Centre, Singapore Management University, Singapore 188065 (email: shawntian@smu.edu.sg)}
\thanks{C. Zhang is with Department of Industrial Engineering, Tsinghua University, Beijing 100084, China (email: zhangchen01@tsinghua.edu.cn)}
\thanks{B. Zheng is with School of Information Systems, Singapore Management University, Singapore 188065 (email: bhzheng@smu.edu.sg)}}

\maketitle


\begin{abstract}
The metro system is playing an increasingly important role in the urban public transit network, transferring a massive human flow across space everyday in the city. In recent years, extensive research studies have been conducted to improve the service quality of metro systems. Among them, crowd management has been a critical issue for both public transport agencies and train operators. In this paper, by utilizing accumulated smart card data, we propose a statistical model to predict in-situ passenger density, i.e., number of on-board passengers between any two neighbouring stations, inside a closed metro system. The proposed model performs two main tasks: i) forecasting time-dependent Origin-Destination (OD) matrix by applying mature statistical models; and ii) estimating the travel time cost required by different parts of the metro network via truncated normal mixture distributions with Expectation-Maximization (EM) algorithm. Based on the prediction results, we are able to provide accurate prediction of in-situ passenger density for a future time point. A case study using real smart card data in Singapore Mass Rapid Transit (MRT) system demonstrate the efficacy and efficiency of our proposed method. 
\end{abstract}

\begin{IEEEkeywords}
Expectation–Maximization algorithm, metro systems, passenger crowding prediction, origin-destination matrix, smart card data, truncated normal distribution
\end{IEEEkeywords}

\section{Introduction}
\label{sec:introduction}
\vspace{-0.05in}

Metro, as one of the most efficient public transport modes, has been an important component in urban development for land-scarce and rail-rely countries like Japan and Singapore. However, as the urban population grows, the increasing public transport demands, especially during peak hours, brings up concern regarding both passenger safety and  metro operation security. In such cases, real-time prediction of passenger density inside the metro network is highly demanded for both system operators and passengers. Accurate and fine-grain prediction can support metro agencies for better train operation planning, assist to detect potential abnormal traffic flow and render fast remedial strategies, and provide real-time traffic information to passengers for better travel planning and overcrowding avoidance. 

Nowadays, the automated fare collection system provides accessibility of massive metro trip data (a.k.a., smart card data) for public transport studies. With the AFC system, when passengers tap their cards on an entry/exit card reader, information such as date, time, location (station ID), and travel direction are recorded. Such data can be used for study of passenger flow distribution, origin-destination (OD) matrix, travel time distribution, trip purposes, 
passenger route choice preferences,
travel behavior analysis,
etc.~\cite{pelletier2011smart} presents a comprehensive review of uses of smart card data in public transit. 

Though there exist some works performing passenger flow prediction by utilizing AFC data, most of them were conducted at the aggregate level, e.g., estimation of passenger inflow and outflow at each station, or the passenger flow of each OD pair. Yet methods providing finer-grain in-situ passenger density estimation across the whole metro network, i.e., number of on-board passengers in each rail segment between two neighboring stations, are not available. This is due to the fact that metro is a closed system, and AFC data only captures trip information of boarding/alighting stations and timestamps, without tracking passengers' real-time locations inside the metro system. Furthermore, as most metro systems are designed to be fault tolerant, there could be multiple routes linking an origin station to a destination station. However the actual route taken by an individual passenger remains unknown, which brings additional challenges to prediction of in-situ passenger density.

To address the above mentioned issues, in this research work, we propose a statistical inference framework, namely \emph{PIPE}, to provide accurate and fine-grain prediction of in-situ passenger density across the metro network, by utilizing collected AFC data. PIPE involves two tasks: a) forecasting time-dependent OD matrices; This aims to infer the potential number of passengers in each OD path; and b) inferring travel time variability of each transit link (to be defined in Section~\ref{sec:preliminary}) of the metro network. This aims to infer how the passengers inferred by a) distribute in different segments of the network. In particular, we directly apply existing machine learning algorithms for the first task. The core of the framework is the second task.
%
Here we propose a white-box statistical inference model by assuming that the travel time distribution for each transit link of the metro system follows a truncated Gaussian distribution. We infer the distribution parameters together with the population-level route choice probabilities using truncated Gaussian mixture models. By leveraging the inferences of both tasks, we can achieve accurate prediction of in-situ passenger density inside the metro system. 

Note that PIPE is different from the in-train passenger load prediction models~\cite{heydenrijk2018supervised,vandewiele2017predicting,pasini2019lstm}, which generally require a lot of additional information, such as on-board headcount data, train timetables or passenger GPS data as input. Yet PIPE only requires AFC data. 
Furthermore, compared with in-train crowding prediction, PIPE aims to infer the macroscopic spatiotemporal passenger density inside the network, rather than the route or train each individual passenger chooses or boards.
In conclusion, PIPE's contribution is twofold: 
\begin{itemize}
\item[(a)] PIPE represents the first attempt to make fine-grain prediction of in-situ passenger density in each rail segment across a closed metro system.
\item[(b)] By taking the metro network topological structure into consideration, PIPE proposes a truncated Gaussian mixture model to infer the travel time of each rail segment of the metro system, and to analyze the population-level route choice preferences, which brings high interpretability to the modelling result. 
\end{itemize}

The remainder of this paper is organized as follows. Section~\ref{sec:literature_review} reviews existing works related to this paper. Section~\ref{sec:preliminary} presents the preliminaries of our work, including trip reconstruction process and generation of route choice sets. Section~\ref{sec:methodology} details the proposed PIPE framework. 
Section~\ref{sec:case_study} reports a case study using Singapore metro AFC data.
Finally, Section~\ref{sec:conclusion} provides concluding remarks.

\section{Literature Review}
\label{sec:literature_review}
\vspace{-0.05in}
Some of the most related research topics include i) passenger flow prediction, ii) OD matrix prediction,  iii) travel time estimation and route selection, and iv) spatiotemporal crowding analysis.


\vspace{0.03in}
\noindent\textbf{Passenger Flow Prediction.}
Passenger flow prediction has been extensively studied in many literature works, among which statistical and machine learning-based approaches have become increasingly attractive. In particular, \cite{Hong2017,xu2018railway,chen2019subway} proposed several time series models, such as autoregressive integrated moving average (ARIMA) model and state space model, for short-term passenger flow prediction at each metro station. Besides, online decomposition based methods, such as non-negative matrix factorization model~\cite{gong2018network} and wavelet decomposition model~\cite{sun2015novel}, have also been proposed, where the passenger flow features are extracted and used for prediction. 

Recently deep neural network-based methods have also been developed for extracting the complex spatial and temporal dependence structure of traffic flows. Various network structures have been applied for metro passenger flow prediction, such as the most classical multiscale radial basis function networks~\cite{li2017forecasting}, stacked auto-encoder~\cite{lv2014traffic}, Long Short Term Memory (LSTM) model~\cite{liu2019deeppf}, sequence-to-sequence model with attention mechanism~\cite{hao2019sequence}, etc. However, all the models developed so far only predict passenger counts, i.e., the inflow and outflow, at the station level, regardless of where the passengers come from or where they are headed, let alone how they are distributed in the system.

\vspace{0.03in}
\noindent\textbf{OD Matrix Forecasting.}
OD matrix forecasting aims at predicting trip demands between different nodes (metro stations in our case) of the network. 
Some pioneer works use linear statistical models for analysis. For example,~\cite{ashok2002estimation} used Kalman filters to predict time-dependent OD flows of drivers, based on traffic volume and average speed data collected using on-board sensors.
A similar approach was also proposed in~\cite{chen2011short} for short-term forecasting of OD Matrix for bus boarding in China.~\cite{van2012dynamic} uses ARIMA model for dynamic forcasting of time-dependent OD flows in a dutch passenger rail. One crucial limitation of the aforementioned linear models is that they cannot represent complex relationship that non-linear models do. 

Recently, newly developed deep learning models, e.g., Recurrent Neural Networks (RNN), have also been widely used for sequence prediction problems. 
For example, \cite{toque2016forecasting} applies LSTM to predict OD matrix in the metro system. ~\cite{hu2020stochastic} proposed a matrix factorization embedded graph CNN for city ground transportation OD matrics prediction. ~\cite{shi2020predicting} proposes a Multi-Perspective
Graph Convolutional Networks (MPGCN) with LSTM to extract temporal features for OD matrix prediction. Yet these methods treat the metro system as a black box and only predict the enter/exit of passengers, but cannot track the passenger locations or infer the in-situ passenger density.

\vspace{0.03in}
\noindent\textbf{Travel Time Estimation and Route Selection.}
To better track passengers' trajectory inside the metro system, many recent researches rely on statistical modelling
to infer travel time variability and passenger route choices. 

These models generally characterize trip travel time between each OD pair as mixture distributions from its candidate routes. Travel time of each candidate route can then be decomposed into time of each transit link, such as the platform waiting time, in-vehicle travel time, transfer time, etc. Passengers' route selection can be influenced by a variety of factors such as the number of transfers required and the total travel time. Different assumptions on the travel time, such as constant~\cite{zhao2016estimation}, Gaussian distribution~\cite{Sun2015,xu2018learning,tian2020tripdecoder}, Poisson distribution~\cite{NIPS2017}, have also been discussed in the literature. However, some of them require additional knowledge, such as the train schedule table~\cite{zhao2016estimation,xu2018learning} and train crowding information~\cite{Sun2015}, which is not always available in reality. Furthermore, all these methods aimed at recovering the routes taken by individual passengers, and did not provide a solution to make macroscopic forecasting of in-situ passenger density across the metro system.

\vspace{0.03in}
\noindent\textbf{Spatiotemporal Crowding Analysis.}
To our best knowledge, only a few works utilize AFC data for crowdedness estimation or passenger distribution inference inside metro systems. In particular, \cite{sun2012using} constructed a regression model to extract the spatial distribution of passengers by dividing them into two groups, based on whether they are travelling on the train or waiting at the platform. However, this method focused on a single track scenario that is oversimplified. \cite{zhang2015spatiotemporal,zhao2016estimation} proposed an empirical probability model to estimate the route choice probabilities from the perspective of individual passenger based on the AFC data and train operating time table, and further extracted spatio-temporal segmentation information of trips as a by-product. As an alternative, some models also aim at directly estimating the in-train passenger density. For example, \cite{pasini2019lstm} constructed a LSTM encoder-predictor combined with a contextual representation for train load prediction. \cite{jenelius2019data} also proposed Boosted Regression Tree Ensemble for both train-centered prediction and station-centered crowdedness prediction. All the above models require additional information, such as the train operating timetable and passenger load of each train car. However, these types of information are not always available, which hinder their applications in general cases.  


\section{Preliminary}
\label{sec:preliminary}

In this section, we first propose a trip reconstruction process in Subsection~\ref{subsec:trip_reconstruction}, which decomposes a trip into a sequence of travel steps. We formulate the metro system as an undirected network and generate a feasible route set for each OD pair of the metro network in Subsection~\ref{subsec:route_set_generation}. These two steps lay foundation for in-situ passenger density prediction. 

\subsection{Trip Reconstruction}
\label{subsec:trip_reconstruction}

\begin{figure*}[t!]
\centering
\includegraphics[width=18cm]{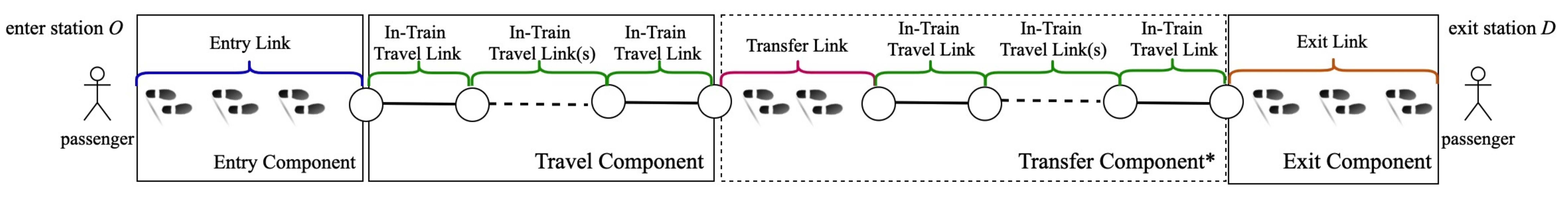} 
\vspace{-0.2in}
\caption{Transit links of a trip in a metro system}
\label{fig:illustration_path} 
\vspace{-0.1in}
\end{figure*}
We model a metro network as a general transportation graph $G(S, E, L)$, consisting of a set of metro stations $S$, a set of edges $E$, and a set of metro lines $L$. A station $s \in S$ could be either a normal station that is crossed by only one metro line or an interchange station that is crossed by multiple lines. An edge (or a segment, interchangeably) $e(s_i,s_j,l) \in E$ is defined as a segment on a train line $l \in L$ that connects the two neighbouring stations $s_i$ and $s_j$ without passing any other station. Stations $s_i$ and $s_j$ are adjacent if there is an edge $e(s_i,s_j,l) \in E$ between them. Note that there could be multiple edges between two adjacent stations $(s_i, s_j)$, corresponding to different metro lines. This undirected network formulation is reasonable since most metro systems in the world are bi-directional. However, the techniques developed in this paper could be easily extended to support the case where a metro system has single-directional lines and should be modelled as a directed graph. 

A route $r_{ij}$ from an origin station $s_i$ to a destination station $s_j$ is a sequence of adjacent edges $\langle e_1, \cdots, e_{L_{ij}^{r_{ij}}}\rangle$ that could bring passengers from station $s_i$ to station $s_j$. 
In this paper, we only consider simple routes without loop, so that each route only visits a station at most once. 

We denote $T_{ij}^{r_{ij}}$ as the corresponding travel time required when a passenger takes a particular route $r_{ij}$ to travel from the origin station $s_i$ to the destination station $s_j$ (note there could be multiple possible routes which will be detailed later). As illustrated in Fig.~\ref{fig:illustration_path}, a trip normally consists of three components: the \emph{entry component}, the \emph{travel component}, and the \emph{exit component}. If the route taken requires transfers, an additional \emph{transfer component} is involved. Accordingly, we can model $T_{ij}^{r_{ij}}$ by decomposing it into travel time of the following four kinds of travel components described above.  
\begin{itemize}
  \item $T_{s_i}^{g}$ represents the time required by an entry link, consisting of the walking time from an entry turnstile at the origin station $s_i$ to the platform and the waiting time for next train at the platform. 
  \item $T_{e}^{c}$ represents the time required by a travel link, consisting of time spent in travelling on edge $e$;
  \item $T_{s}^{q}$ represents the time required by a transfer link, consisting of the walking time from one metro platform to another, and the waiting time for the next train at an interchange station $s$; and
  \item $T_{s_j}^a$ represents the time required by an exit link, i.e., the walking time from the platform to the turnstiles at the destination station $s_j$.
\end{itemize}
Hereafter the term \emph{transit link} is used to refer to one component of a trip via a metro system, which contributes to the total time required by a trip from entering the origin station to exiting the destination station.

Given an edge $e(s_i,s_j,l_x)$ connecting station $s_i$ and station $s_j$ along service line $l_x$, we assume the travel time required from $s_i$ to $s_j$ via service line $l_x$ is identically distributed as that required from $s_j$ to $s_i$ via the same line. This assumption generally holds for most metro systems. Yet our analytic framework could be easily extended to cases when the travel time from $s_i$ to $s_j$ is asymmetric, even along the same service line.

After decomposing a trip into four different types of transit links, we can sum up the time spent on each transit link of $r_{ij}$ and calculate the total travel time required by $r_{ij}$ as stated in Equation~(\ref{equ:Tr}).

\begin{equation}
T_{r_{ij}} = T_{s_i}^{g}+\sum\nolimits_{b=1}^{L_{ij}^{r_{ij}}}{T_{e_b}^{c}}+\sum\nolimits_{s\in S_{ij}^{r_{ij}}}{T_s^{q}}+T_{s_j}^a.
\label{equ:Tr}
\end{equation}
Here $S_{ij}^{r_{ij}}$ refers to the set of interchange stations on route $r_{ij}$ where passengers make transfers. If we can derive the travel time required by each transit link involved in $r_{ij}$, we can predict the exit time of a trip, given its entry time. In addition, we can also infer the position of a passenger in the metro system at any time point before (s)he ends the trip.
\vspace{-0.2in}
\subsection{Route Choice Set Generation}
\label{subsec:route_set_generation}
Commonly, in a metro system, there could be multiple routes for some OD pairs.
In the following, the term \emph{route choice set} corresponding to each OD pair $\langle i, j\rangle$, denoted as $R_{ij}$, represents all the routes used by passengers to travel from $s_i$ to $s_j$. We could adopt different strategies to generate $R_{ij}$, such as edge elimination and $k$-shortest-paths. 
%
%
In this paper, considering the number of stations in a metro system is usually in the scale of either tens or hundreds (e.g., as the largest metro system in the world, New York City Subway has in total 400+ stations). We simply adopt brute-force-search algorithm to form $R_{ij}$ for different OD pairs. 

In the search process, note that NOT all the available routes are actually practical, e.g., passengers do not prefer a route that is much longer or with too more transfers than others. Therefore we exclude routes that satisfy at least one of the following criteria from $R_{ij}$: i) routes with any loops; ii) routes that are not the shortest path (in terms of number of transit links) but require more than $\sigma$ transfers; and iii) routes with $\beta$ ($>1$) times number of links than the shortest route $r_{ij}^{min}$, i.e., the route with a minimum number of transit links out of $R_{od}$. 
%
%
The controlling parameters $\beta$ and $\sigma$ could be set according to the assumptions of passengers' behavior. For example, in our study, we set both $\beta$ and $\sigma$ to be two. The notation $M_{ij}$ stands for the number of routes inside the route choice set $R_{ij}$.
\section{In-SITU PASSENGER DENSITY PREDICTION}
\label{sec:methodology}

\begin{figure}[t!]
\centering
\includegraphics[width=0.48\textwidth]{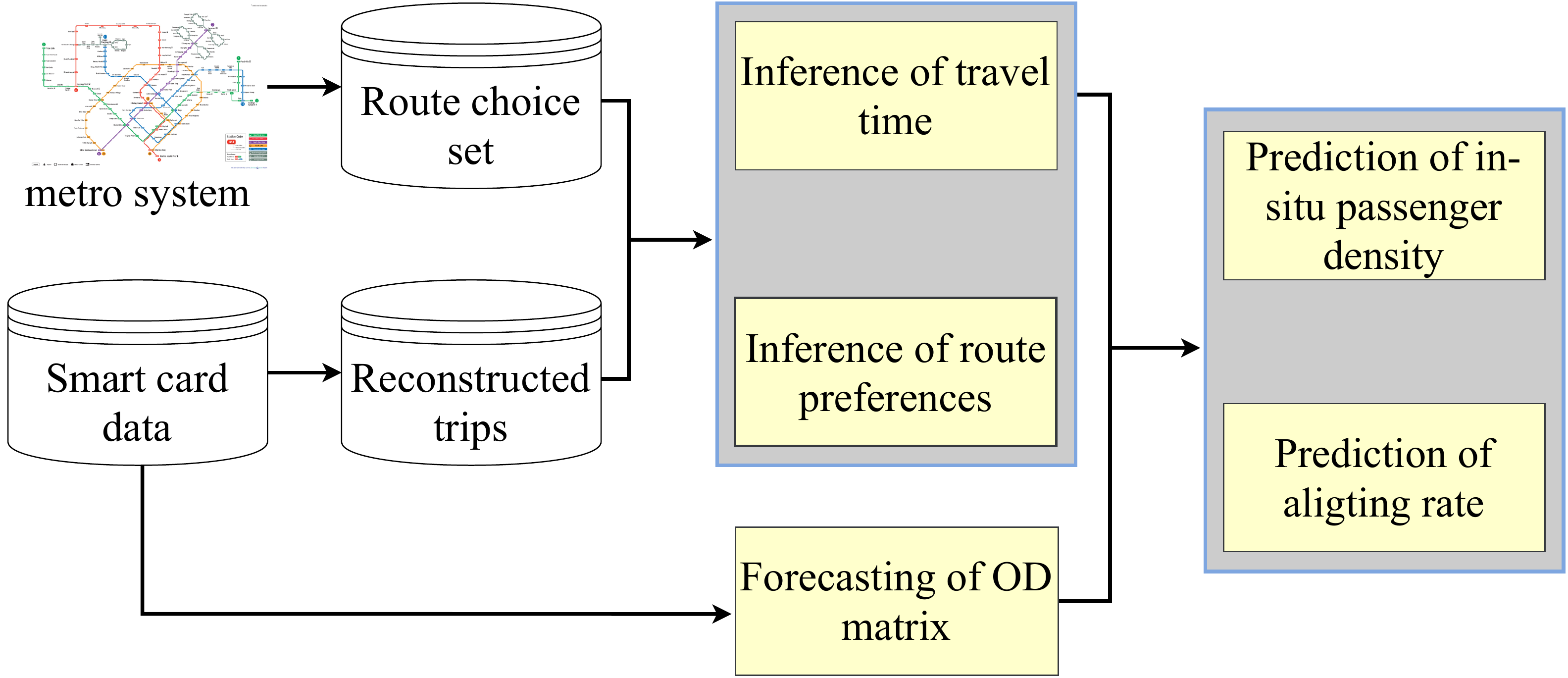}
\caption{\textsc{PIPE} framework for \underline{p}redicting the metro crowding of \underline{i}n-situ \underline{p}ass\underline{e}ngers.}
\vspace{-0.1in}
\label{fig:model_framework}
\end{figure}

In this section, we propose a framework namely \emph{PIPE} to \underline{p}redict the \underline{i}n-situ \underline{p}assenger d\underline{e}nsity in the metro system based on AFC data. 
%
%

Given a metro system $G(S,E,L)$, for a particular day, we would like to predict the passenger density on edge $e$ at some point $t$ in the future, i.e., $X_e(t)$. This can be formulated as:
\begin{equation}
\begin{split}
\label{eq:in-train}
X_e(t) = \sum_{i \in S}\sum_{j \neq i}\int_{\tau<t}V_{ij}(\tau)P_{ij|e}P(T_{i,e}=t-\tau) d \tau
\end{split}
\end{equation}
where 
\begin{itemize}
\item $V_{ij}(\tau)$: the number of passengers boarding at $s_i$ at time $\tau$ and later alighting at $s_j$,
\item $P_{ij|e}$: the probability that a passenger boarding at $s_i$ and alighting at $s_j$ would take a route containing edge $e$, 
\item $P(T_{i,e}=t-\tau)$: the probability that it takes time $t-\tau$ for a passenger boarding at $s_i$ to reach edge $e$.  
\end{itemize}

Similarly, we can model the number of passengers alighting at $s_j$ in future time $t$, i.e., $X_j(t)$, as:
\begin{equation}
\begin{split}
\label{eq:tap_out_flow}
X_{j}(t) = \sum\nolimits_{i \in S \land i \neq j}\int_{\tau<t}V_{ij}(\tau)P(T_{ij}=t-\tau) d \tau
\end{split}
\end{equation}
where 
\begin{itemize}
\item $P(T_{ij}=t-\tau)$: the probability that it takes time $t-\tau$ for a passenger to travel from $s_i$ to $s_j$.
\end{itemize}

Without loss of generality, we assume the AFC data set $\calD$ including in total $N$ historical metro trips, i.e., $\calD=\cup_{n=1}^N tr_n$. Each metro trip is represented as $tr=(id,o,d,t_o,t_d,c)$, where $id$ is an encrypted unique string identifying a smart card, $o$ is the origin station, $d$ is the destination station, $t_o$ and $t_d$ record the timestamp when the passenger enters the station $o$ and exits the station $d$ respectively, and $c\in C$ refers to the passenger category (e.g., $C = \{$child, adult, senior, student$\}$ in Singapore). $T=t_d - t_o$ captures the real travel time required by this trip. For brevity, we also represent a trip as $tr=(id,o,d,t_o,T,c)$.
Based on $\calD$, we want to predict $X_e(t)$ and $X_{j}(t)$. To achieve the prediction goals, we need to infer $V_{ij}(\tau)$, the distributions of $T_{i,e}$ and $T_{ij}$, and $P_{ij|e}$.
These actually can be divided into two tasks in PIPE: i)  forecasting the number of passengers travelling between each OD pair given the entry time window $\tau$\footnote{In this paper, we break time into 20-minute time windows}, i.e., $V_{ij}(\tau)$; and ii) estimating the travel time parameters for all transit links and route choice probabilities for all routes in $R_{ij}$. Fig.~\ref{fig:model_framework} plots the architecture of the proposed solution, we detail each of its components in subsequent subsections. 

\subsection{Prediction of $V_{ij}(\tau)$}
\label{subsec:estimation_od_matrix}
As introduced in Section \ref{sec:literature_review}, many works have been devoted to predict $V_{ij}(\tau)$ for a particular day (here without confusion, we omit the day subscription for brevity) using statistical or machine learning techniques in the last decade. In this paper, we consider the following candidates. 

Define $X_{k,i}^{in}(t)$, $X_{k,i}^{out}(t)$, $V_{k,ij}(t)$ as the passenger inflow and outflow of $s_{i}$, the passenger flow from $s_{i}$ to $s_{j}$ at time window $t$ in day $k$ of the training data set. Assume we have $k=1,\ldots,K$ days of data.  
The first vanilla candidate is a calendar model using the historical average of the $K$ days to predict OD matrix for the testing day, 
, i.e., predicting $V_{ij}(\tau)$ for a given time window $\tau$ by averaging trip counts that occurred in that same time window of the $K$ days, i.e., $V_{ij}(\tau) = \sum_{k=1}^{K}V_{k,ij}(\tau)/K$.

Besides the vanilla method, we explore the following machine learning and deep learning algorithms:

\begin{itemize}
\item \emph{Linear regression models}: We consider
\begin{align}
V_{ij}(\tau) &=\sum_{s\in S}\sum_{l=1}^{\Delta}\left[ a_{ij}^{s,l}X_{s}^{in}(\tau-l)+b_{ij}^{s,l}X_{s}^{out}(\tau-l) \right ] \\ \nonumber
\end{align}
where $a_{ij}^{s,l}$, $b_{ij}^{s,l}$ are regression coefficients of passenger inflow and outflow at station $s$ with a lag order of $l$ respectively, $\Delta$ is the maximum lag order decided based on validation performance. Considering the number of inputs is high, we introduce regularization to the coefficients by using Lasso~\cite{tibshirani1996regression} and Ridge~\cite{hoerl1970ridge}) to filter out unrelated inputs. 
\item \emph{Random forest model}~\cite{liaw2002classification}: Each decision tree takes lagged passenger inflow $X_{s}^{in}(t-l)$ and outflow $X_{s}^{out}(t-l), l=1, \ldots, \Delta$, as predictors, the final prediction is the mean predictions of all individual trees.
\item The time series \emph{ARIMA model}~\cite{makridakis1997arma}: We formulate the problem as an univariate time series prediction problem. Trip counts of each OD pair $V_{k,ij}(t)$ are sorted by date and time in ascending order and use ARIMA for model fitting. In particular, we adopt a walk-forward validation method to evaluate model performance. It is a practice used to evaluate time series models when the model is expected to be updated sequentially as new observations are available. For each time window in the testing day, a model constructed based on the training dataset will be used for prediction. Then the observation of the current time window will be added to the training dataset and the process repeat. When performing multi-step ahead prediction, say $m$-step ahead, the prediction results from the past windows $V_{ij}(\tau+1),\ldots,V_{ij}(\tau+m-1)$, instead of the ground truth values, are taken as observations for predicting $V_{ij}(\tau+m)$.
\item \emph{LSTM}~\cite{hochreiter1997long}: The OD matrix of the 100 past time windows $\tau-l, l=1,\ldots, 100$ are taken as inputs to predict $V_{ij}(\tau)$. For the training step, we used a validation-based early stopping~\cite{prechelt1998early}. This method allows to avoid overfitting by stopping the training of the model when the loss of the validation set stops decreasing. We use the ADAM implementation of Stochastic gradient descent (SGD) for weights optimization. The model architecture consist a LSTM layer with $9,000$ hidden units and a fully connected layer. 
\end{itemize}
For ARIMA and LSTM, it is to be noted that in our case for some long traval time routes, when predicting $V_{ij}(\tau)$, $V_{ij}(\tau-1)$, $V_{ij}(\tau-2)$,...,$V_{ij}(\tau-d)$ are possibly unknown. This is because passengers of $V_{ij}(\tau-d)$ can still be in the middle of trip and have not tapped out yet. Here $d$ is the order of lag depending on the travel duration form station $i$ to station $j$. For this case, we simply remove these $d$ windows' data from the input set. Take Singapore MRT network as an example, as almost all trips can be finished within two hours except that a MRT break down happened, we can set $d=6$ (, i.e., 2 hours) whenever forecasting $V_{ij}(\tau)$. For example, when performing one-step ahead prediction of $V_{ij}(\tau)$, only $V_{ij}(\tau-7)$, $V_{ij}(\tau-8)$,... are considered as inputs.

\subsection{Estimation of $P(T_{i,e})$, $P(T_{i,j})$ and $P_{ij|e}$}
\label{subsec:estimation_travel_time}

For a certain OD pair $\left<i,j\right>$, we assume i) it has $m=1,\ldots,M_{ij}$ possible routes and ii) the route $m$ includes $L_{ij}^{m}$ edges.
Then, the total travel time of a trip taking route $m$ can be decomposed into the travel time of different transit links, as stated in Equation~(\ref{equ:t_ij}). For notation convenience, we denote all these transit links as set $\calH_{ij}^{m}$, and the total travel time equals the sum of the time required by each transit link in $\calH_{ij}^{m}$:

\begin{equation}
\begin{split}
T_{ij}^{m} = T_{s_i}^{g} + \sum_{b=1}^{L_{ij}^{m}}T_{e_b}^{c} + \sum_{s\in \calS_{ij}^{m}} T_{s}^{q} + T_{s_j}^{a}
= \sum_{h\in \calH_{ij}^{m}}T_{h}.
\end{split}
\label{equ:t_ij}
\end{equation}

In order to consider travel time variability, we assume travel time $T^{g}_{s}$, $T^{c}_{e}$, $T^{q}_{s}$ and $T^{a}_{s}$ follow truncated Gaussian distributions~\cite{wang2012speed}, i.e., $T_{s}^{g} \sim TN(\mu_{s}^{g},\sigma_{s}^{g},a_{s}^{g},b_{s}^{g})$,  $T_{e}^{c} \sim TN(\mu_{e}^{c}$,$\sigma_{e}^{c},a_{e}^{c},b_{e}^{c})$, $T_{s}^{q} \sim TN(\mu_{s}^{q},\sigma_{s}^{q},a_{s}^{q},b_{s}^{q})$, and $T_{s}^{a} \sim TN(\mu_{s}^{a},\sigma_{s}^{a},a_{s}^{a},b_{s}^{a})$, 
where the probability distribution function of the truncated Gaussian distribution $TN(\mu,\sigma,a,b)$ is defined as 
%
\begin{align}
x\sim TN(\mu,\sigma,a,b) = \left\{
\begin{array}{ll}
\frac{\phi(\frac{x-\mu}{\sigma})}{\sigma(\Phi(\frac{b-\mu}{\sigma})-\Phi(\frac{a-\mu}{\sigma}))} & \text{if } a\leq x \leq b,\\
0 & \text{otherwise }.
\end{array} \right. \nonumber
\end{align}

Then, the distribution of $T_{ij}^{m}$ can be also approximated by a truncated Gaussian distribution~\cite{Cozman-1994-13767} as
\begin{align}
T_{ij}^{m}& \sim TN(\mu_{ij}^{m},\sigma_{ij}^{m},a_{ij}^{m},b_{ij}^{m}),
\end{align}
where
\begin{align}
\mu_{ij}^{m}& = \mu_{s_i}^{g}+\sum_{b=1}^{L_{ij}^{m}}\mu_{e_b}^{c} + \sum_{s\in \calS_{ij}^{m}} \mu_{s}^{q} +\mu_{s_{j}}^{a}=\sum_{h \in \calH_{ij}^{m}}\mu_{h}\\
\sigma_{ij}^{m^2}  &= \sigma_{s_i}^{g^2} + \sum_{b=1}^{L_{ij}^{m}}\sigma_{e_b}^{c^2} + \sum_{s\in \calS_{ij}^{m}} \sigma_{s}^{q^2} +\sigma_{s_{j}}^{a^2} =\sum_{h \in \calH_{ij}^{m}}\sigma_{h}^{2}\\
a_{ij}^{m}  &=   a_{s_i}^{g}+\sum_{b=1}^{L_{ij}^{m}}a_{e_b}^{c} + \sum_{s\in \calS_{ij}^{m}}a_{s}^{q}+ a_{s_{j}}^{a} =\sum_{h \in \calH_{ij}^{m}}a_{h}\\
b_{ij}^{m}  &= b_{s_i}^{g}+\sum_{b=1}^{L_{ij}^{m}}b_{e_b}^{c}+ \sum_{s\in \calS_{ij}^{m}} b_{s}^{q}+b_{s_{j}}^{a}=\sum_{h \in \calH_{ij}^{m}}b_{h}
\end{align}

For OD pairs $\left<i,j\right>$ with single possible route, i.e., $M_{ij}=1$, the travel time $T_{ij} \sim TN(\mu_{ij}^{1},\sigma_{ij}^{1},a_{ij}^{1},b_{ij}^{1})$. For OD pair $\left<i,j\right>$ with multiple possible routes, i.e., $M_{ij}>1$, we assume $T_{ij}$ follows a truncated Gaussian mixture distributions: 
\begin{equation}
\begin{split}
T_{ij} \sim \sum\nolimits_{m=1}^{M_{ij}}\pi_{ij}^{m,c}TN(\mu_{ij}^{m},\sigma_{ij}^{m},a_{ij}^{m},b_{ij}^{m}).
\end{split}
\end{equation}
Here, $\pi_{ij}^{m,c}$ is the probability that passengers choose route $r_{ij}^m$ out of $R_{ij}$, and $\sum_{m=1}^{M_{ij}}\pi_{ij}^{m,c}=1$, with $c$ representing the category of passengers. For example, in Singapore, there are four categories of passengers, i.e., $C=\{Adult, Child, Senior, Student\}$. Based on our observation, route preferences could differ between passenger categories. For example, seniors may prefer more comfort routes which requires longer travel time but are less crowded, since old people are more flexible in terms of time and yet are physically more vulnerable. In contrast, commuters, who generally rush for time, probably prefer the shortest routes even though they are super crowded.

Now we discuss about how to estimate the above truncated Gaussian mixture models. In particular, we use parameter set $\bTheta$ to represent $\{\mu_{s}^{g},\sigma_{s}^{g}\}$, $\{\mu_{s}^{q},\sigma_{s}^{q}\}$, $\{\mu_{s}^{a},\sigma_{s}^{a}\}$ for any station $s \in S$, $\{\mu_{e}^{c},\sigma_{e}^{c}\}$ for any edge $e \in E$, and $\bpi_{ij}=\{\pi_{ij}^{1,c},\ldots,\pi_{ij}^{M_{ij},c};\forall M_{ij}>1,\forall c \in C\}$. We use maximum likelihood estimation to estimate $\bTheta$ based on accumulated AFC data $\calD = \{tr_{n}(id,o,d,t_{o},T,c),n=1,\dots,N\}$. As for each trip in $\calD$, if its OD pair has more than one possible route, the route choice information is missing. We propose to use Expectation-Maximization (EM) method to estimate the route choice preferences together with $\btheta$. In particular, suppose the missed route choice information for trip $tr_{n}$ is known as $\Z_{n}$. Here, if only one route is available, $Z_{n}=1$; if in total $M_{n}$ routes are available, $\Z_{n}=[Z_{n1},\ldots,Z_{n M_n}]$. $Z_{nm}=1$ if the route taken by $tr_n$ is $r_{o_n d_n}^m$ and $Z_{nm}=0$ otherwise. Then, we have $\calZ=\{\Z_{n},n=1,\ldots,N\}$. Consequently, the full likelihood of a particular $\bTheta$ given
the AFC card dataset $\calD$ and route choice $\calZ$ can be formulated as
\begin{equation}
\begin{split}
\label{eq:likelihood}
& \calL \left(\bTheta|\calD,\calZ \right) = \prod\nolimits_{n=1}^{N} \Big[I_{M_{o_nd_n}=1}TN\left( t_n|\mu_{n}^1,\sigma_{n}^1,a_{n}^1,b_{n}^1 \right) \\
& +I_{M_{o_n d_n}>1}\prod_{m=1}^{M_{o_n d_n}} \Big( \pi_{o_n d_n}^{m,c_n} TN \left( t_{n}|\mu_{n}^{m},\sigma_{n}^{m},a_{n}^{m},b_{n}^{m}\right)\Big)^{Z_{nm}} \Big].
\end{split}
\end{equation}
For label convenience, we abuse the notation $\mu_{n}^{m} = \mu_{o_nd_n}^{m} = \sum_{h \in \calH_{o_n d_n}^{m}}\mu_{h}$, $\sigma_{n}^{m^2} = \sigma_{o_nd_n}^{m^2} =  \sum_{h \in \calH_{o_n d_n}^{m}}\sigma_{h}^{2}$, $a_{n}^{m}=a_{o_n d_n}^{m}=\sum_{h \in \calH_{o_n d_n}}a_{h}$, and $b_{n}^{m}=b_{o_n d_n}^{m}=\sum_{h \in \calH_{o_n d_n}^{m}}b_{h}$. Taking the logarithm of Equation~\eqref{eq:likelihood}, we can get the log-likelihood as 
\begin{equation}
\begin{split}
\label{eq:log_likelihood}
 l(\bTheta|\calD,\calZ)  =& \sum\nolimits_{n=1}^{N} \Big\{I_{M_{o_n d_n}=1}\ln\Big(TN\left(t_{n}|\mu_{n},\sigma_{n},a_{n},b_{n}\right)\Big) \\
+& I_{M_{o_n d_n}>1}\sum\nolimits_{m=1}^{M}Z_{nm}\Big[\ln(\pi_{o_n d_n}^{m,c_n}) \\
+& \ln \Big(TN\left(t_{n}|\mu_{n}^{m},\sigma_{n}^{m},a_{n}^{m},b_{n}^{m}\right)\Big)\Big]\Big\}.
\end{split}
\end{equation}

However, in reality, the route information $Z_{nm}$ is unknown and hence Equation~(\ref{eq:log_likelihood}) cannot be solved directly. The idea of EM algorithm is to iteratively estimate $\bTheta^{(k+1)}$ by maximizing the expectation of the complete log-likelihood function, i.e.,  $E_{\calZ|\calD,\bTheta^{(k)}}\left[l\left(\bTheta|\calZ,\calD\right)\right]$, given the current estimated parameters $\bTheta^{(k)}$. In our formulation, this can be achieved via replacing $Z_{nm}$ by $E\left(Z_{nm}|\calD,\bTheta^{(k)}\right)$ in Equation~ \eqref{eq:log_likelihood}. In particular,
\begin{equation}
\begin{split}
\tilde{Z}_{nm} =& E\left(Z_{nm}|\calD,\bTheta^{(k)}\right) \\
=& \frac{\pi_{o_n d_n}^{m,c_n (k)}TN\left(t_{n}|\mu_{n}^{m(k)},\sigma_{n}^{m(k)},a_{n}^{m(k)},b_{n}^{m(k)}\right)}{\sum_{m=1}^{M_{o_nd_n}}\pi_{o_n d_n}^{m,c_n (k)}TN\left(t_{n}|\mu_{n}^{m(k)},\sigma_{n}^{m(k)},a_{n}^{m(k)},b_{n}^{m(k)}\right)}
\end{split}
\end{equation}

Then, we have
\begin{equation}
\begin{split}
\label{eq:expectation}
& E_{\calZ|\calD,\bTheta^{k}}\left[l\left(\bTheta|\calZ,\calD\right)\right] = \tilde{l}\left(\calD,\tilde{\calZ}|\bTheta\right)\\
=& \sum_{n=1}^{N} \Big\{ I_{M_{o_nd_n}=1}\ln\Big(TN\left(t_{n}|\mu_{n},\sigma_{n},a_{n},b_{n}\right)\Big)\\
+& I_{M_{o_n d_n}>1}\sum_{m=1}^{M}\tilde{Z}_{nm}\Big[\ln\left(\pi_{o_n d_n}^{m,c_n}\right)\\
+& \ln \Big(TN\left(t_{n}|\mu_{n}^{m},\sigma_{n}^{m},a_{n}^{m},b_{n}^{m}\right)\Big)\Big]\Big\}.
\end{split}
\end{equation}

We reformulate the parameters and update $\{\mu_{s}^{g},\sigma_{s}^{g^2}\}$, $\{\mu_{s}^{a},\sigma_{s}^{a^2}\}$, $\{\mu_{s}^{q},\sigma_{s}^{q^2}\}$ for $s \in S$, $\{\mu_{e}^{c},\sigma_{e}^{c^2}\}$ for $e\in E$, and $\bpi_{ij}^{m,c}=\{\pi_{ij}^{1,c},\ldots,\pi_{ij}^{M_{ij},c};\forall M_{ij}>1, \forall c \in C\}$ separately. For example, to maximize $\{\mu_{s}^{g}, \sigma_{s}^{g}\}$, we extract the part of Equation~\eqref{eq:expectation} that relates to $\{\mu_{s}^{g}, \sigma_{s}^{g^2}\}$ as:
\begin{equation}
\begin{split}
\label{eq:expectation_example}
\tilde{l}(\mu_{s}^{g},\sigma_{s}^{g^2})
=& \sum_{n=1}^{N} \Big\{I_{(M_{o_n d_n}=1,o_n=s)} \ln\Big(TN(t_{n}|\mu_{n},\sigma_{n},a_{n},b_{n})\Big)\\
+& I_{(M_{o_n d_n}>1,o_n=s)}  \sum_{m=1}^{M}\tilde{Z}_{nm}\Big[\ln(\pi_{o_nd_n}^{m,c_n})\\
+& \ln \Big(TN\left(t_n|\mu_{n}^{m},\sigma_{n}^{m},a_{n}^{m},b_{n}^{m}\right)\Big)\Big]\Big\}
\end{split}
\end{equation}
The maximization of Equation (\ref{eq:expectation_example}) has no closed form solution. Consequently we apply stochastic gradient descent (SGD) to update the value of $\{\mu_{s}^{g}, \sigma_{s}^{g^2}\}$ in a iterative way. In particular, the first derivatives of (\ref{eq:expectation}) with respect to $\mu_{s}^{g}$ and $\sigma_{s}^{g^2}$ are $\G=[\frac{\partial \tilde{l}}{\partial \mu_{s}^{g}},\frac{\partial \tilde{l}}{\partial \sigma_{s}^{g^2}} ]$ with

 \begin{equation}
 \begin{split}
 \label{eq:1st_der}
 \frac{\partial \tilde{l}}{\partial \mu_{s}^{g}} =&\sum_{n=1}^{N} I_{(M_{o_n d_n}=1,o_n=s)}\left[\frac{1}{\sigma_{n}}\frac{\phi(\frac{b_{n}-\mu_{n}}{\sigma_{n}})-\phi(\frac{a_{n}-\mu_{n}}{\sigma_{n}})}{\Phi(\frac{b_{n}-\mu_{n}}{\sigma_{n}})-\Phi(\frac{a_{n}-\mu_{n}}{\sigma_{n}})} \right.\\ 
 & \left.+\frac{(Y_{n}-\mu_{n})}{\sigma_{n}^{2}} \right] 
+\sum_{n=1}^{N}I_{(M_{o_n d_n}>1,o_n=s)}  \left[\sum_{m=1}^{M}\tilde{Z}_{lm} \right.\\ &\left.\left(\frac{1}{\sigma_{n}^{m}} \frac{\phi(\frac{b_{n}^{m}-\mu_{n}^{m}}{\sigma_{n}^{m}})-\phi(\frac{a_{n}^{m}-\mu_{n}^{m}}{\sigma_{n}^{m}})}{\Phi(\frac{b_{n}^{m}-\mu_{n}^{m}}{\sigma_{n}^{m}})-\Phi(\frac{a_{n}^{m}-\mu_{n}^{m}}{\sigma_{n}^{m}})}+\frac{(Y_{n}-\mu_{n}^{m})}{\sigma_{n}^{m^2}} \right) \right],\\
 \end{split}
 \end{equation}
 
 \begin{equation}
 \begin{split}
 \frac{\partial \tilde{l}}{\partial \sigma_{s}^{g^2}} =&\sum_{n=1}^{N}I_{(M_{o_n d_n}>1,o_n=s)} \left[\sum_{m=1}^{M}\tilde{Z}_{nm}\left(
\frac{(Y_{n}-\mu_{n}^m)^2}{2\sigma_{n}^{m^4}}-\frac{1}{2\sigma_{n}^{m^2}} \right.\right.\\ 
&\left.\left.+ \frac{1}{2\sigma_{n}^{m^3}}\frac{(b_{n}^{m}-\mu_{n}^{m})\phi(\frac{b_{n}^{m}-\mu_{n}^{m}}{\sigma_{n}^{m}})-(a_n^m-\mu_{n}^m)\phi(\frac{a_{n}^m-\mu_{n}^m}{\sigma_{n}^{m}})}{\Phi(\frac{b_{n}^m-\mu_{n}^m}{\sigma_{n}^{m}})-\Phi(\frac{a_{n}^m-\mu_{n}^m}{\sigma_{n}^{m}})} \right)\right] \\
&+\sum_{n=1}^{N}I_{(M_{o_n d_n}=1,o_n=s)} \left[\frac{(Y_{n}-\mu_{n})^2}{2\sigma_{n}^{4}}-\frac{1}{2\sigma_{n}^{2}} \right.\\ 
 &\left.+\frac{1}{2\sigma_{n}^{3}}\frac{(b_{n}-\mu_{n})\phi(\frac{b_{n}-\mu_{n}}{\sigma_{n}})-(a_n-\mu_{n})\phi(\frac{a_{n}-\mu_{n}}{\sigma_{n}})}{\Phi(\frac{b_{n}-\mu_{n}}{\sigma_{n}})-\Phi(\frac{a_{n}-\mu_{n}}{\sigma_{n}})}\right].
 \end{split}
 \end{equation}

Similarly, we can estimate $\{\mu_{s}^{g},\sigma_{s}^{g^2}\}$, $\{\mu_{e}^{c},\sigma_{e}^{c^2}\}$, $\{\mu_{s}^{q},\sigma_{s}^{q^2}\}$ and $\{\mu_{s}^{a},\sigma_{s}^{a^2}\}$ for all transit links.

Thereafter, the updated $\bpi_{ij}^{m,c}$ becomes
\begin{equation}
\begin{split}
\pi_{ij}^{m,c} = \frac{\sum_{n=1}^{N}I_{(o_n = i,d_n=j, c_n=c)}\tilde{Z}_{nm}}{\sum_{m=1}^{M_{ij}}\sum_{n=1}^{N}I_{(o_n = i,d_n=j,c_n=c)}\tilde{Z}_{nm}}
\end{split}
\end{equation}
Now we talk about how to set the truncation points for different transit links. In specific, truncation points $[a,b]$ for transfer link $T_s^q$ are set as $[0, 2*(w_0+l_0)]$, where $w_0$ is the default transfer walking time\footnote{In the context of Singapore, we set $w_0$ as 2 minutes.}, i.e., the walking time from one metro platform to another, and $l_0$ is the train headway of train service $l$. As to the truncation points for $T_s^g$, $T_s^a$ and $T_e^c$, we estimate their values according to the following iterative algorithm~\ref{alg:Truncation}.

\begin{algorithm}
\KwData{$\calD = \cup_{n=1}^N tr(id_n,o_n,d_n,t_n,c_n)$; length of edge $e_{len}$ for $e \in E$; maximum train travel speed $l_{vmax}$ for $l \in L$}
\KwResult{Estimated $\{a_{s}^{g},b_{s}^{g}\},\{a_{s}^{a},b_{s}^{a}\}$ for $s \in S$, $\{a_{e}^{c},b_{e}^{c}\}$ for $e \in E$}
\textbf{initialization}\\
Initialize $\{a_{s}^{g},b_{s}^{g}\},\{a_{s}^{a},b_{s}^{a}\},\{a_{e}^{c},b_{e}^{c}\}$ as 0\\
 \textbf{Estimation}\\
 \For{$e(s_i, s_j, l) \in E$}{
 $Y \gets \{tr \in \calD | (tr.o = s_i \land tr.d = s_j) \lor  (tr.o = s_j \land tr.d = s_i)\}$\\
 $t_{min} \gets min_{tr \in Y} (tr.t)$, 
 $t_{max} \gets max_{tr \in Y}(tr.t)$\\
 $b_e^c \gets t_{min}$, 
 $a_e^c \gets e_{len} / l_{vmax}$\\
 $b_{s_i}^g \gets max(b_{s_i}^g, t_{max}-t_{min})$\\
 $b_{s_i}^a \gets max(b_{s_i}^a, t_{max}-t_{min})$\\
 $b_{s_j}^g \gets max(b_{s_j}^g, t_{max}-t_{min})$\\
 $b_{s_j}^a \gets max(b_{s_j}^a, t_{max}-t_{min})$\\
}
\caption{Estimation of truncation points}
\label{alg:Truncation} 
\end{algorithm}

It is noted that if more information about each individual station is available, $w_{0}$ can be set differently for different stations. 

\section{Case Study}
\label{sec:case_study}
\vspace{-0.05in}
In this section, we apply the proposed PIPE framework in the context of Singapore Mass Rapid Transit (MRT) system. In the following, we first introduce the dataset used in this study and the data preprocessing steps to remove outlier trip data in Section~\ref{subsec:case_study_data}; we then present the case study results, including prediction results of time-dependent OD matrices, travel time parameters, alighting rate at each MRT station and in-situ passenger density across the metro system in Section~\ref{subsec:case_study_result}, to demonstrate the prediction performance of the proposed PIPE framework.

\subsection{Data Set}
\label{subsec:case_study_data}


\subsubsection{Singapore MRT System}
\label{subsubsec:mrt_system}

\begin{figure}[t!]
\centering
\includegraphics[width=0.48\textwidth]{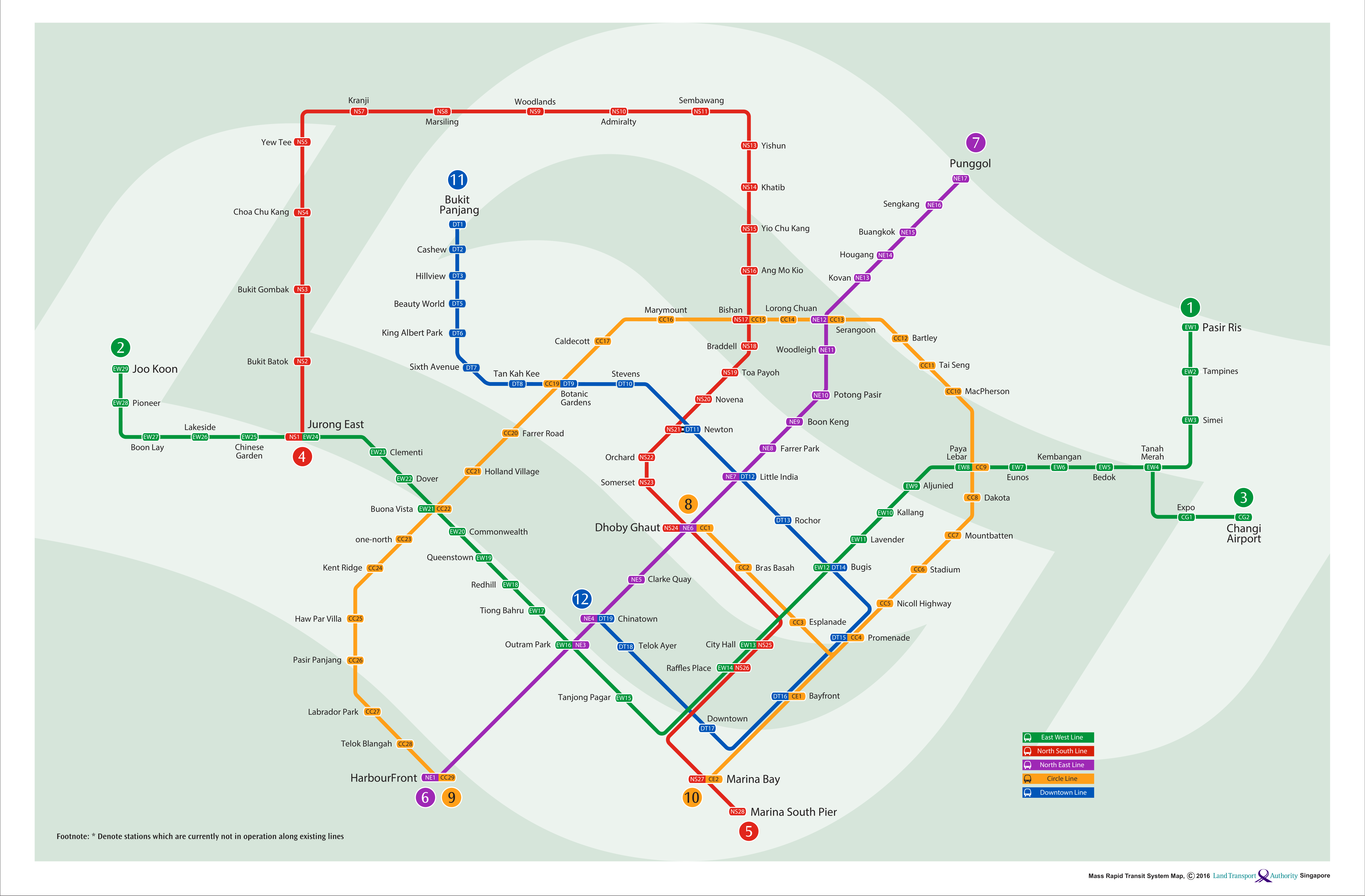}
\caption{Singapore MRT Network Map (as of May 2016)}
\label{fig:mrt_map}
\end{figure}

Singapore's MRT system plays an increasingly important role in Singapore public transit network. Up to May 2016, the MRT network, as shown in Figure~\ref{fig:mrt_map}, consists of $102$ stations, $7$ MRT lines (including two line extensions), and $114$ edges between adjacent stations.

\subsubsection{EZ-Link Card Data}
\label{subsubsec:ezlink_data}

EZ-Link card is the smart card used in Singapore for the payment of public transport trips. 
$251,089,965$ MRT trip records, which were collected from all the working days from January 1 to May 31 in 2016, are utilized as the data source in our study, 
since the passenger flow patterns of working days is significantly different from that of weekends and public holidays. 
However, our framework can be easily applied to include weekend/public holiday data prediction as well. 
%
As introduced in Section \ref{sec:preliminary}, each MRT trip is represented as $Tr(id,o,d,t_o,t_d,c)$ or $Tr(id,o,d,T,c)$, with four sample records listed in Table~\ref{tab:ezlink_sample}.
%

%

\renewcommand{\tabcolsep}{1pt}
\begin{table}[t!]
\centering
\caption{EZ-Link MRT Record Samples}
\label{tab:ezlink_sample}
\begin{tabular}{|p{1.2cm}|p{0.8cm}|p{1.5cm}|p{1.7cm}|p{1.1cm}|p{1.3cm}|}
\hline
\textbf{card id $id$} & \textbf{type $c$}  & \textbf{entry datetime $t_o$}  & \textbf{exit datetime $t_d$}  & \textbf{origin id $o$} &\textbf{destination id $d$}                    
\\ \hline
02***5F & adult & 2016-01-25 08:20:04 & 2016-01-25 08:27:27 & 35 & 12
\\ \hline
02***5F & adult&  2016-01-25 18:13:57 & 2016-01-25 18:21:25 & 12 & 35

\\ \hline
02***5F & adult& 2016-01-26 08:13:51 & 2016-01-26 08:21:21 & 35 & 12
\\ \hline
02***5F & adult& 2016-01-26 18:31:45 & 2016-01-26 18:38:11 & 12 & 35
\\ \hline
\end{tabular}
\vspace{-0.15in}
\end{table}

\subsubsection{Data Pre-Processing}
\label{subsubsec:data_preprocessing}

Due to AFC system deficiency and other technical limitations, some trips are not properly captured. Three types of noisy data are removed before we proceed with the data analysis: i) duplicate records for the same trip; ii) outlier trip records with extremely long travel time identified based on the interquartile range (IQR) rule; and iii) trip records with missing information. Consequently, about $5.3\%$ of the records have been identified as noisy data. 
%
As the size of noisy data is significantly smaller than 
that of the valid data, we assume that the removal of those noisy records will not bias our analysis.

\subsection{Case Study Result and Discussion}
\label{subsec:case_study_result}

In the following, we present the performance of PIPE, including i) OD matrix prediction, ii) travel time distribution inference, iii) alighting rate prediction and iv) in-situ passenger density prediction respectively.
\subsubsection{OD Matrix Prediction}
As presented in Section~\ref{subsec:estimation_od_matrix}, in this paper we consider six different methods to predict OD matrix for a particular time window in a new day, based on previously observed OD matrices and passenger inflow and outflow data of each metro station. 
For each method, we performed a grid search to select the meta parameters, e.g., the number of lags $\Delta$, that can lead to the best results. We divide the five months' EZ-link data into three disjoint datasets as follows: 70\% of the data are used as training set, 20\% are used as the validation set to perform model selection, and the remaining 10\% are used as testing set to evaluate the models. We adopt the \emph{Mean Square Error (MSE)} as the main performance metric. Table \ref{tab:od_matrix_prediction} reports the MSE of the 100 most busy OD pairs, which covers $13.30\%$ of the whole AFC dataset, with different prediction ahead time steps, where the ahead time step refers to the length between current time window and the time window to be predicted.

\renewcommand{\tabcolsep}{1pt}
\begin{table}[t!]
\centering
\caption{OD Matrix Prediction Error (MSE)}
\label{tab:od_matrix_prediction}
\begin{tabular}{|p{4.3cm}|p{1cm}|p{1cm}|p{1cm}|}
\hline
\diagbox[]{Model}{Ahead time} & \textbf{1} & \textbf{4} & \textbf{6}
\\ \hline
Vanilla model & 249.11 & 249.11 & 249.11
\\ \hline
Linear Regression with lasso & 190.83 & 280.23 & 351.28
\\ \hline
Linear regression with ridge & 191.20 & 283.76 & 354.14
\\ \hline
ARIMA & 2136.98 & 2315.65 & 2382.65
\\ \hline
LSTM & 205.21 & 228.00 & 235.10
\\ \hline
Random Forest & \textbf{159.44} & \textbf{182.57} & \textbf{209.86}
\\ \hline
\end{tabular}
\vspace{-0.1in}
\end{table}

As can be observed from the results, among all the models, random forest produces the best results in general, followed by two linear regression models, while ARIMA and LSTM have comparably worse performance. It indicates $V_{ij}(\tau)$ is more related to $X_{s}^{in}(\tau-l), X_{s}^{out}(\tau-l),l=1,\ldots,$ than to $V_{ij}(\tau-l),l=1,\ldots$. This is because in $X_{s}^{in}(\tau-l), X_{s}^{out}(\tau-l),l=1,\ldots$, both temporal relation between the predicted value and the lagged passenger inflow and outflow, and spatial relation between different stations are taken into consideration.
Furthermore, compared with the linear models, random forest can better capture nonlinear relation between $V_{ij}(\tau)$ and $X_{s}^{in}(\tau-l), X_{s}^{out}(\tau-l),l=1,\ldots$. Of course, some other nonlinear spatio-temporal models can be also applied in practice, if better prediction performance can be achieved. Since this part is not the focus of PIPE, in our following analysis, we just select random forest for predicting $V_{ij}(\tau)$, and other methods are left open to practitioners. 
In addition, as we can observe, as the ahead time increases, it has negative influence on all the models, excepted the vanilla model. This is reasonable. As the ahead time step $m$ increase, more accumulated prediction errors in $V_{ij}(\tau+1),\ldots,V_{ij}(\tau+m-1)$ are used as model input, and consequently deteriorate the prediction performance. As to the vanilla model, since it simply utilizes historical average as the prediction, it would not be influenced by the prediction ahead step.

\begin{figure*}[!htb]
    \centering
    \subfigure[Boon Lay to Jurong East]{
        \includegraphics[width=0.3\textwidth]{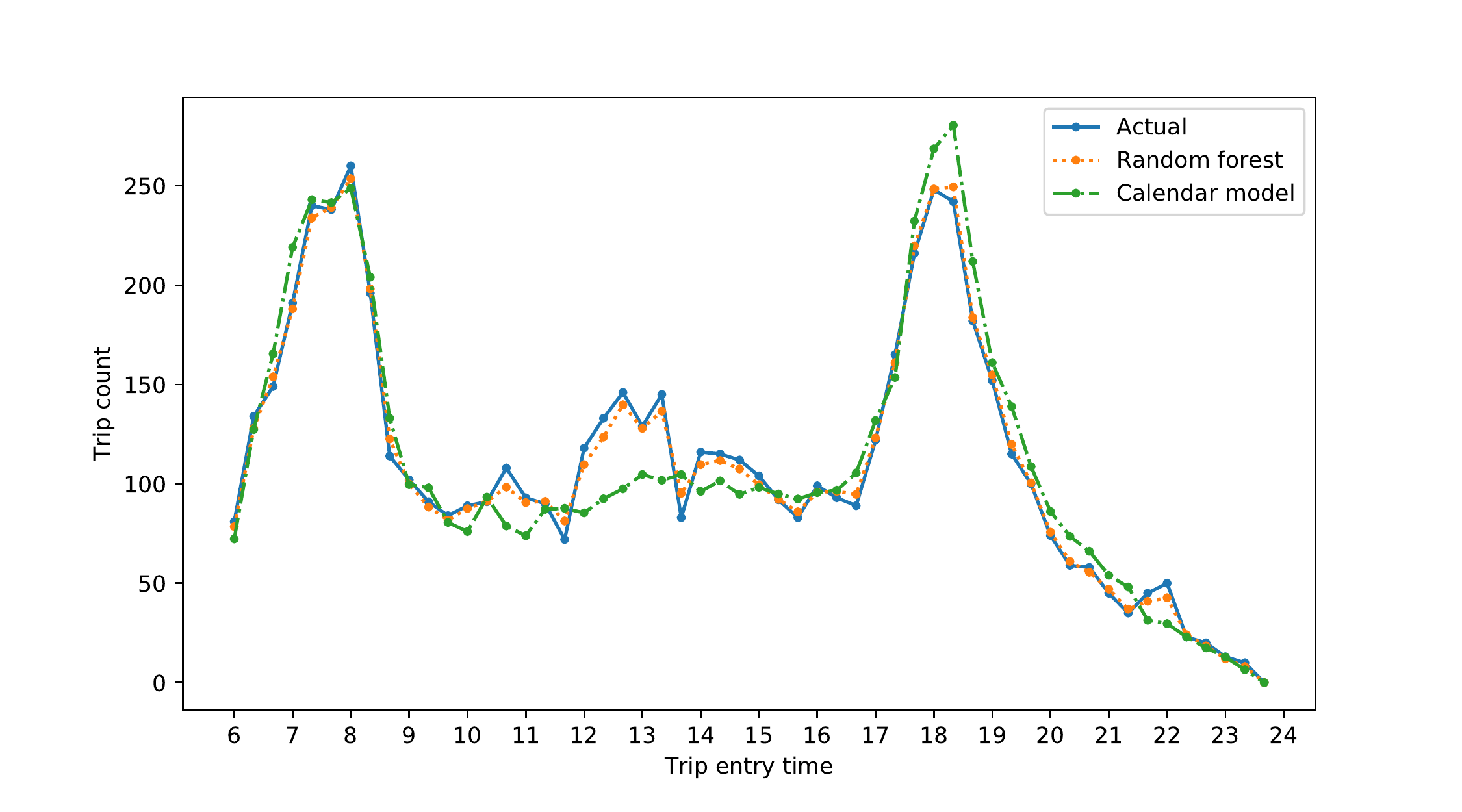} }
    \subfigure[Pasir Ris to Tampines]{
        \includegraphics[width=0.3\textwidth]{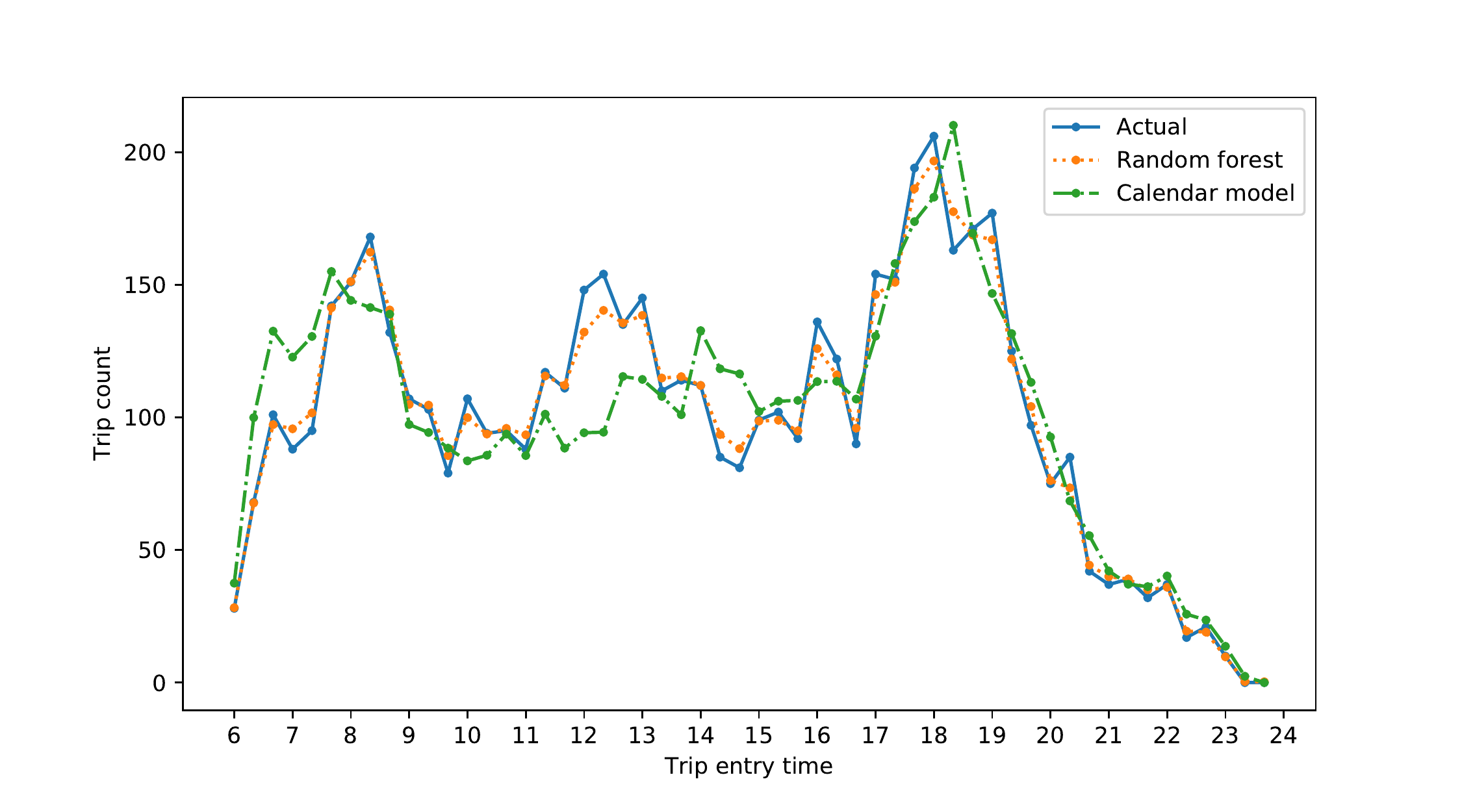} }
    \subfigure[Raffles Place to City Hall]{
        \includegraphics[width=0.3\textwidth]{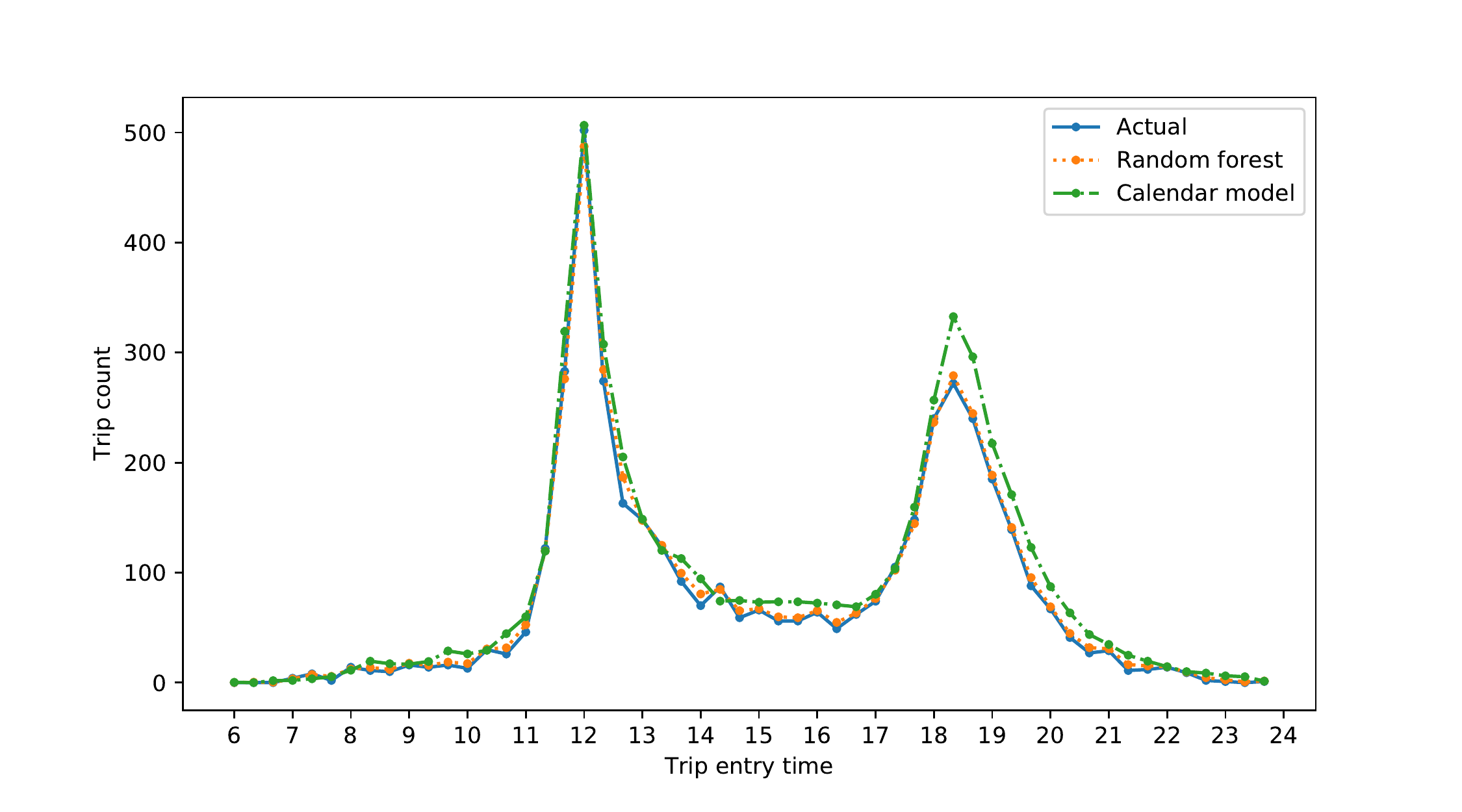} }
    \vspace{-0.1in}
    \caption{OD matrix prediction results}
    \vspace{-0.15in}
\label{fig:od_matrix}
\end{figure*}

To better demonstrate the prediction results, three OD pairs are selected to report the random forest's one-step ahead prediction performance for a particular day in Fig.~\ref{fig:od_matrix}. For all three sample OD pairs, random forest is able to capture small temporal fluctuations of $V_{ij}(\tau)$ much more accurately than the vanilla model. Take $\langle$Boon Lay, Jurong East$\rangle$ in Fig.~\ref{fig:od_matrix} (a) as an example, random forest achieves a MSE of 22.86, which is significantly lower than 334.29 achieved by 
vanilla model.

\subsubsection{Travel Time Prediction}

\begin{figure*}[!htb]
\begin{center}
    \subfigure[Entry link at EW4 ($\mu$=7.03)]{
        \includegraphics[width=0.23\textwidth]{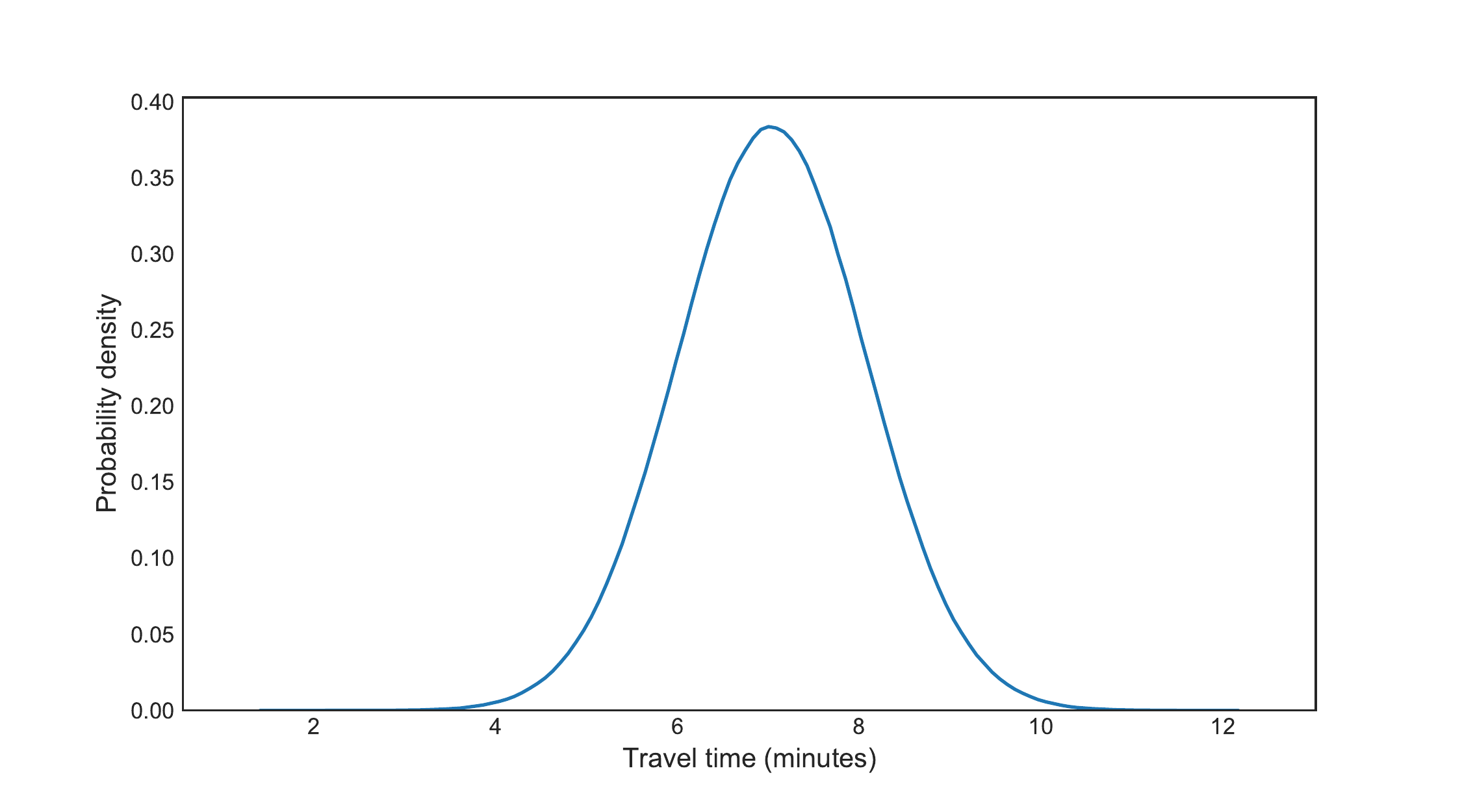}}
    \subfigure[In-train travel link from EW4 to CG1 ($\mu$=1.01)]{
        \includegraphics[width=0.23\textwidth]{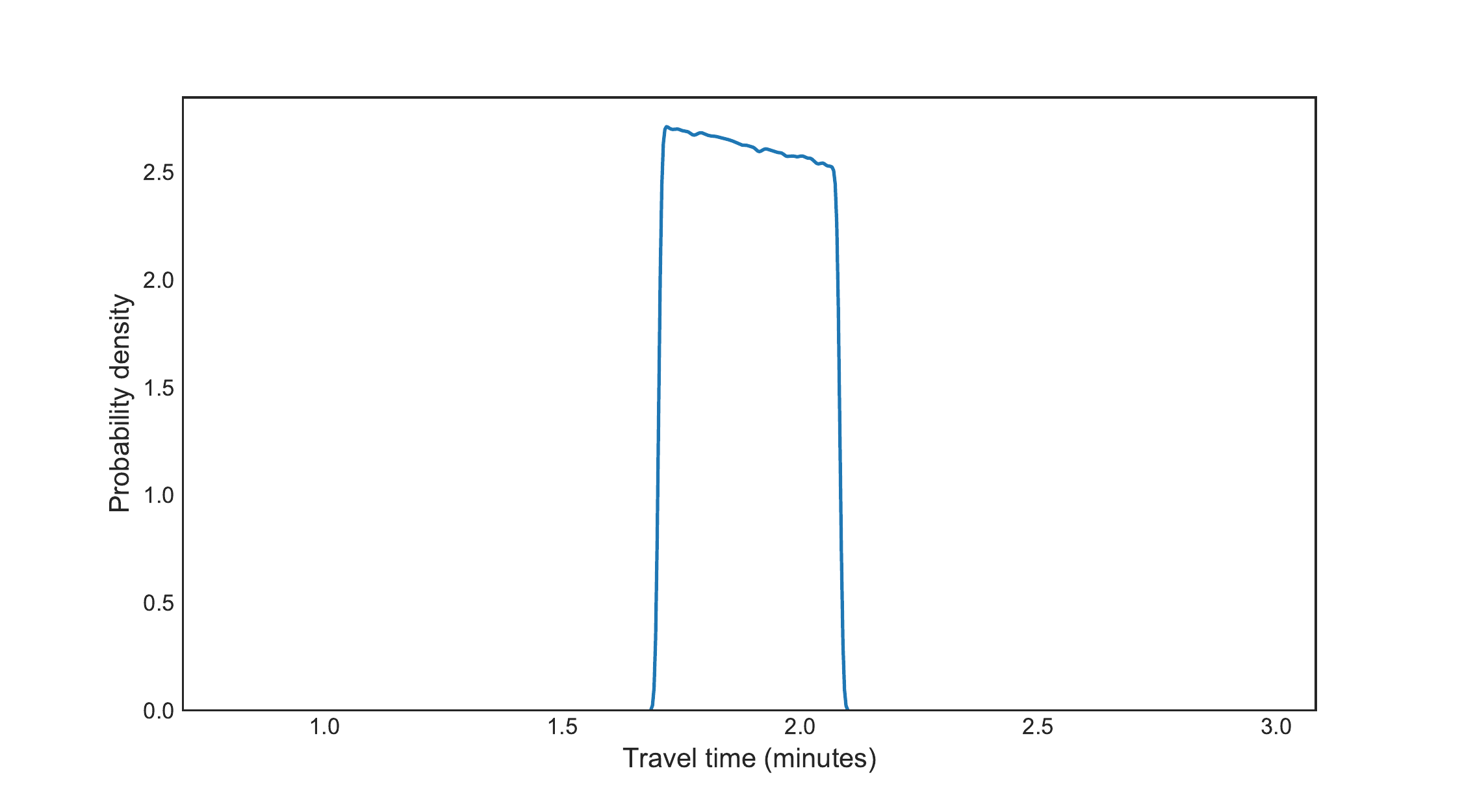}}
    \subfigure[In-train travel link from CG1 to CG2 ($\mu$=4.97)]{
        \includegraphics[width=0.23\textwidth]{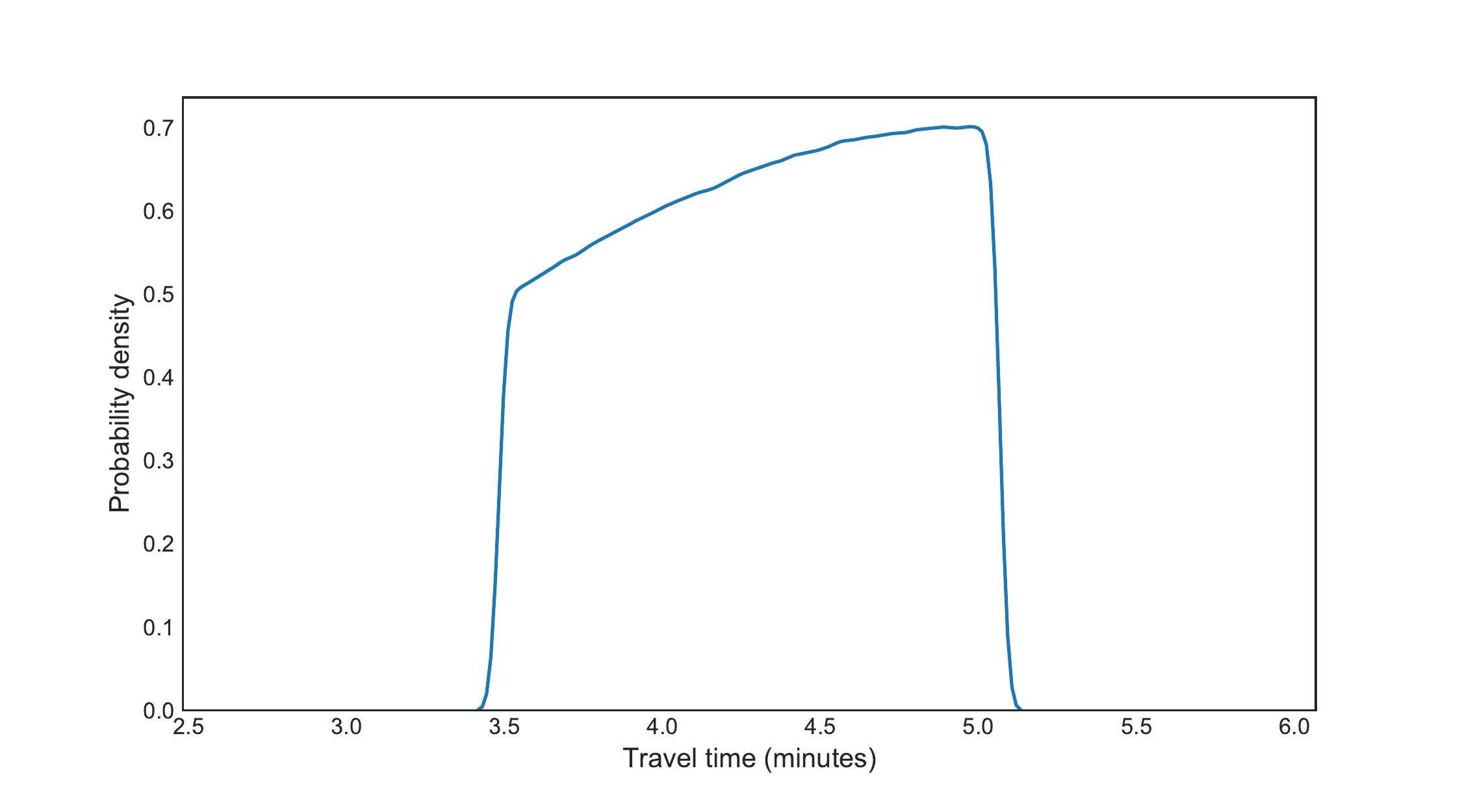}}
    \subfigure[Exit link at CG2 ($\mu$=1.63)]{
        \includegraphics[width=0.23\textwidth]{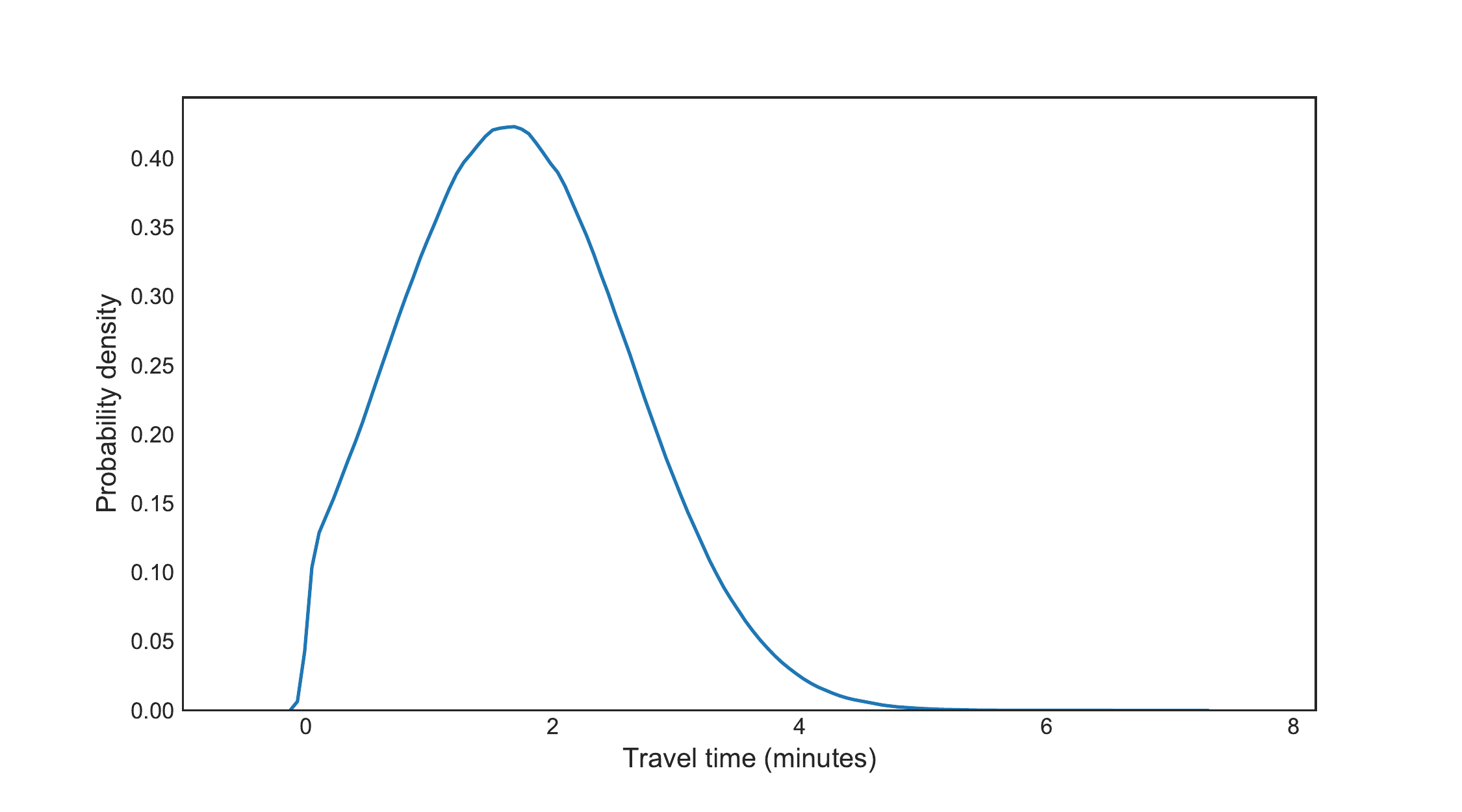}} 
    \vspace{-0.1in}
    \caption{Travel time fit for travel links of sample route (Tanah Merah (EW4) to Changi Airport (CG2))}
    \label{fig:travel_time_fit_link}
    \vspace{-0.15in}
\end{center}
\end{figure*}

\begin{figure*}[!htb]
    \centering
    \subfigure[Admiralty to Boon Lay]{
        \includegraphics[width=0.3\textwidth]{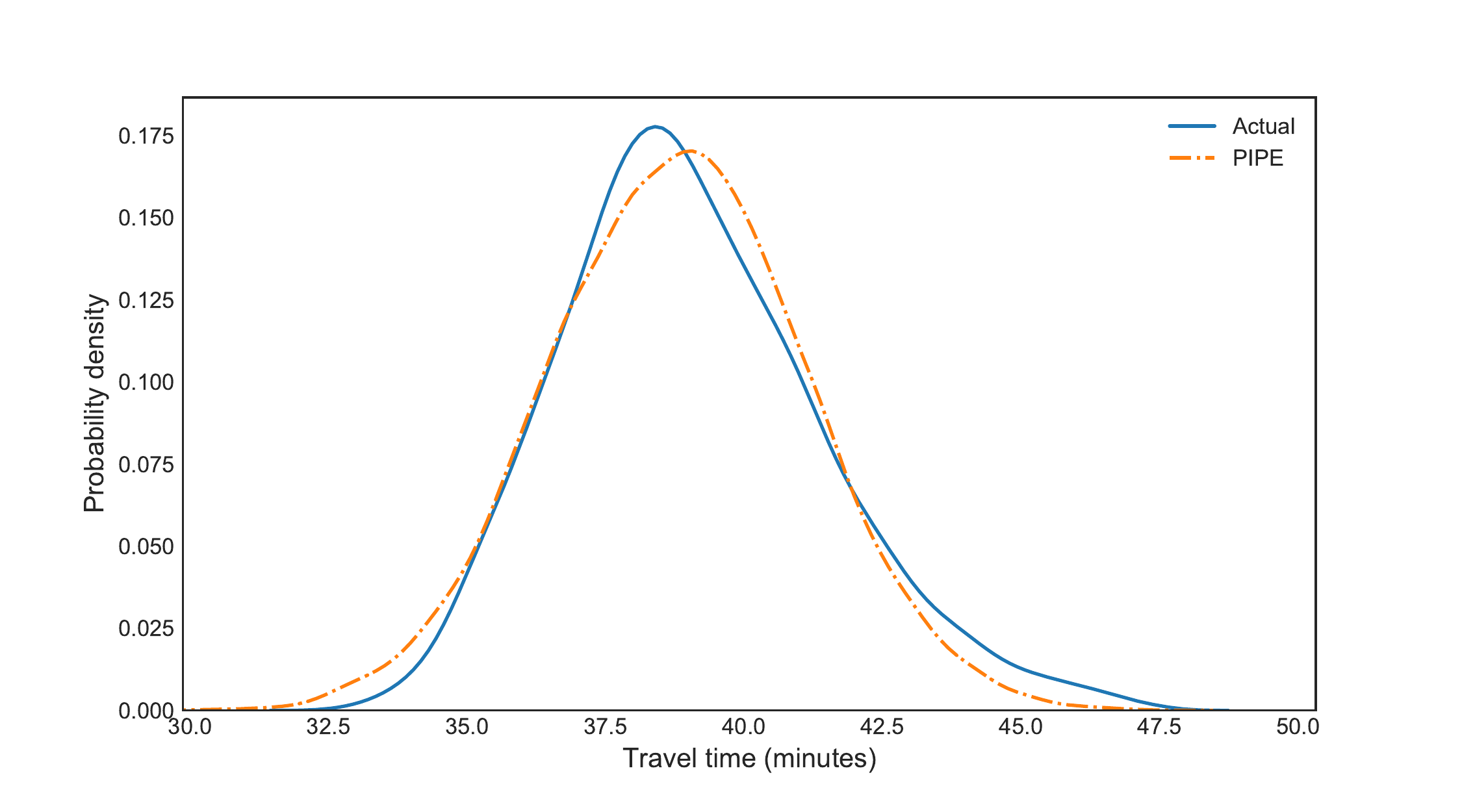}}
    \subfigure[Jurong East to Woodlands]{
        \includegraphics[width=0.3\textwidth]{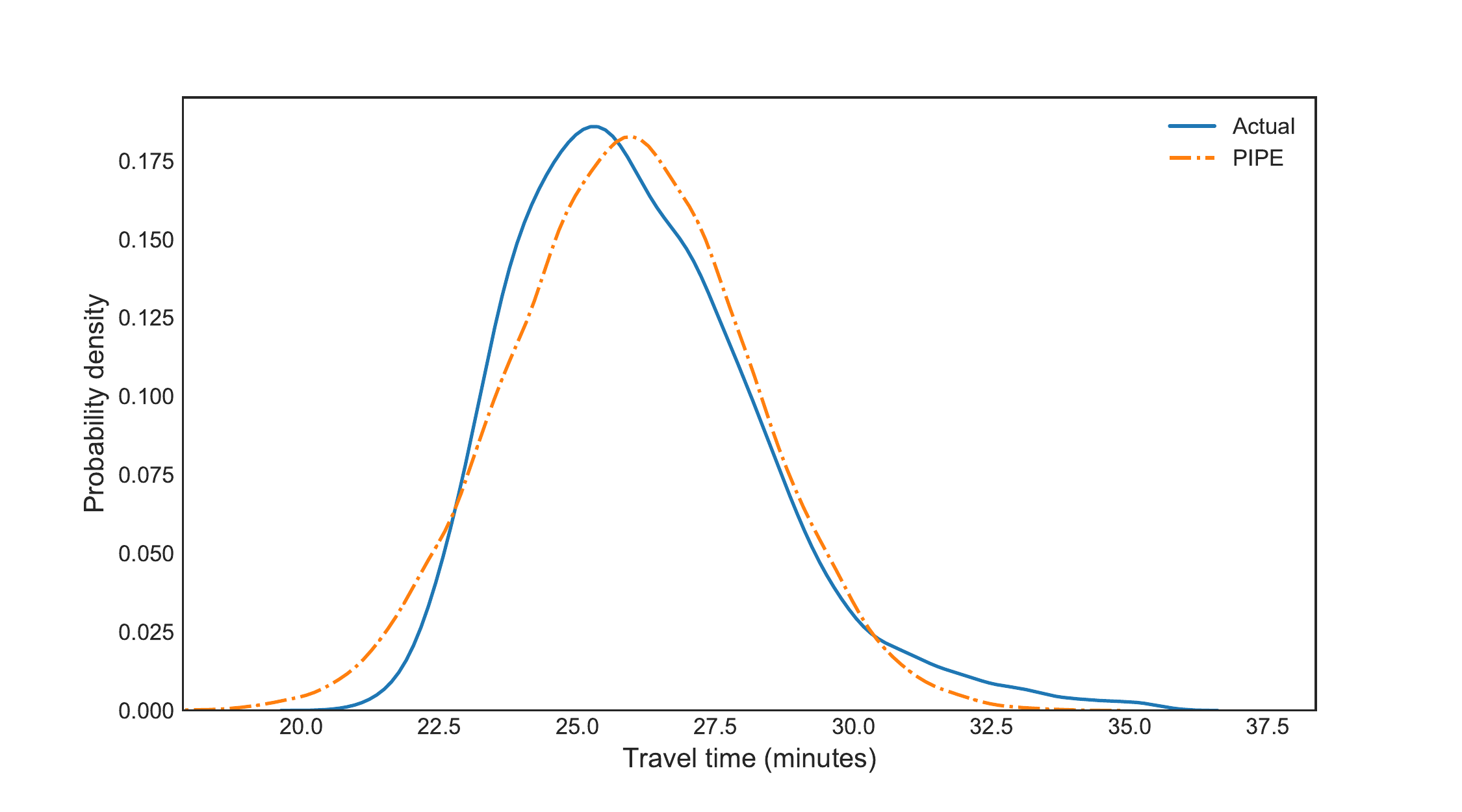} }
    \subfigure[Beauty World to Promenade]{
        \includegraphics[width=0.3\textwidth]{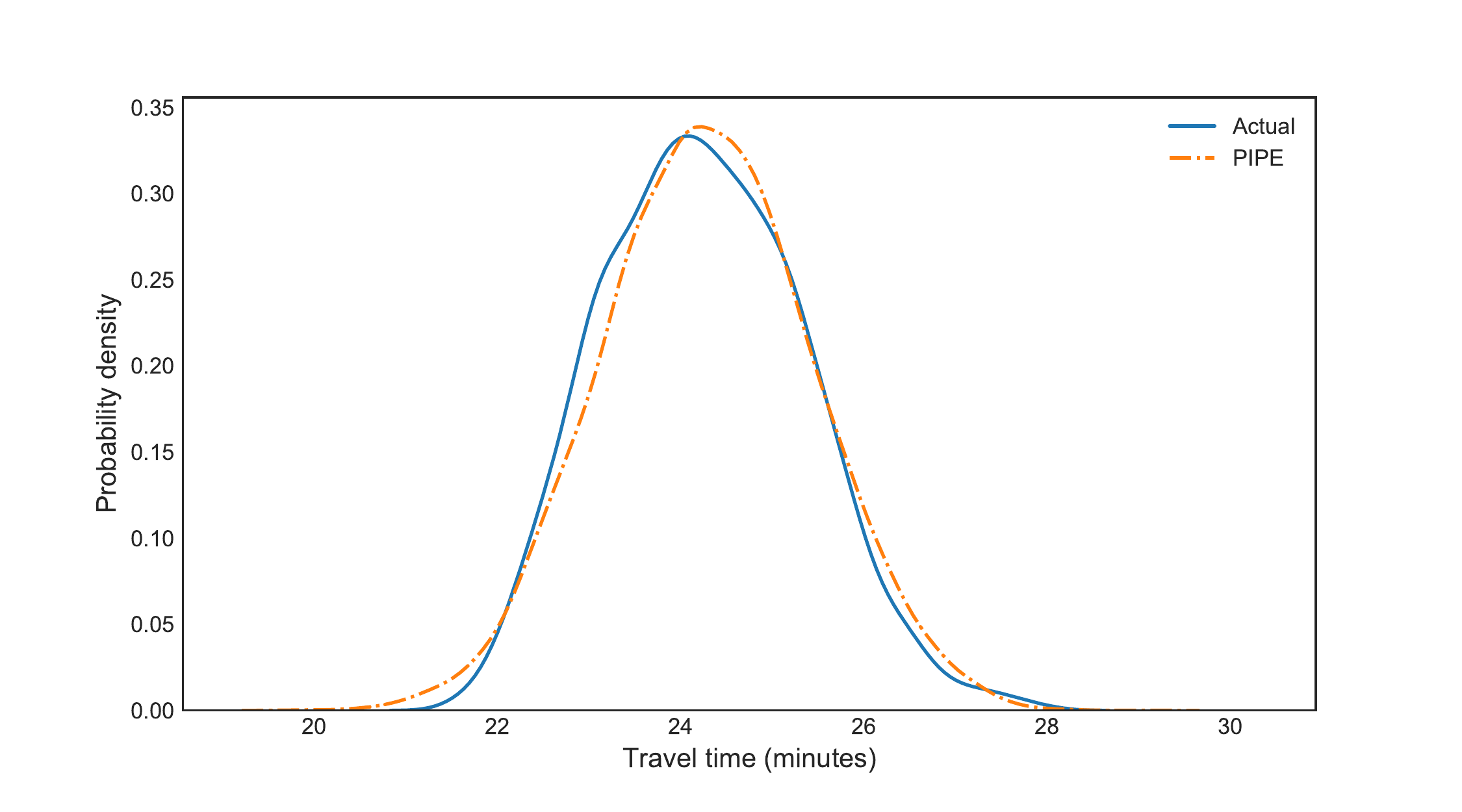} }
    \vspace{-0.1in}
    \caption{Travel time fit for single-route OD pairs}
    \label{fig:travel_time_fit_single}
    \vspace{-0.15in}
\end{figure*}

\begin{figure*}[!htb]
    \centering
    \subfigure[Admiralty to HarbourFront]{
        \includegraphics[width=0.3\textwidth]{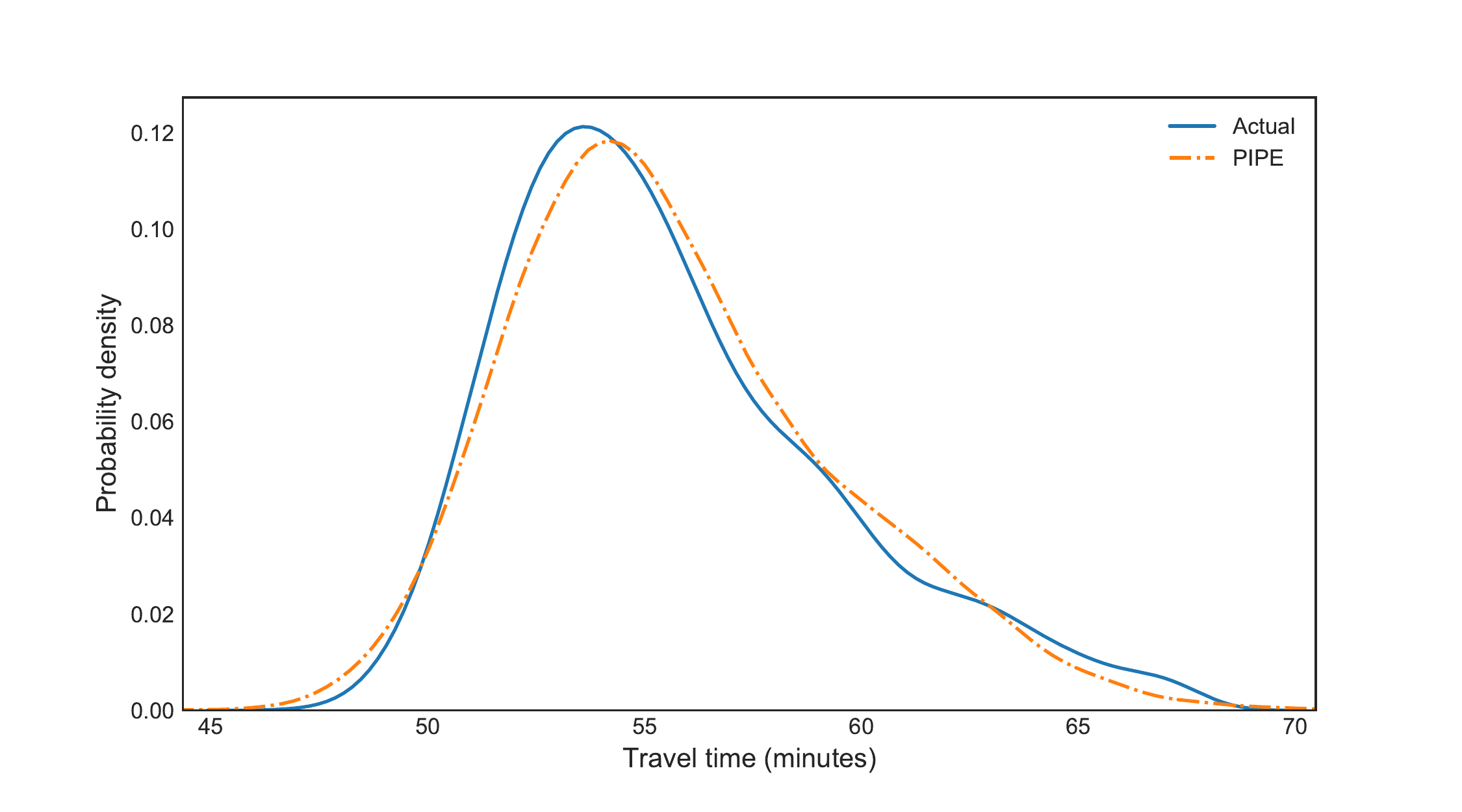} }
    \subfigure[Hougang to Kent Ridge]{
        \includegraphics[width=0.3\textwidth]{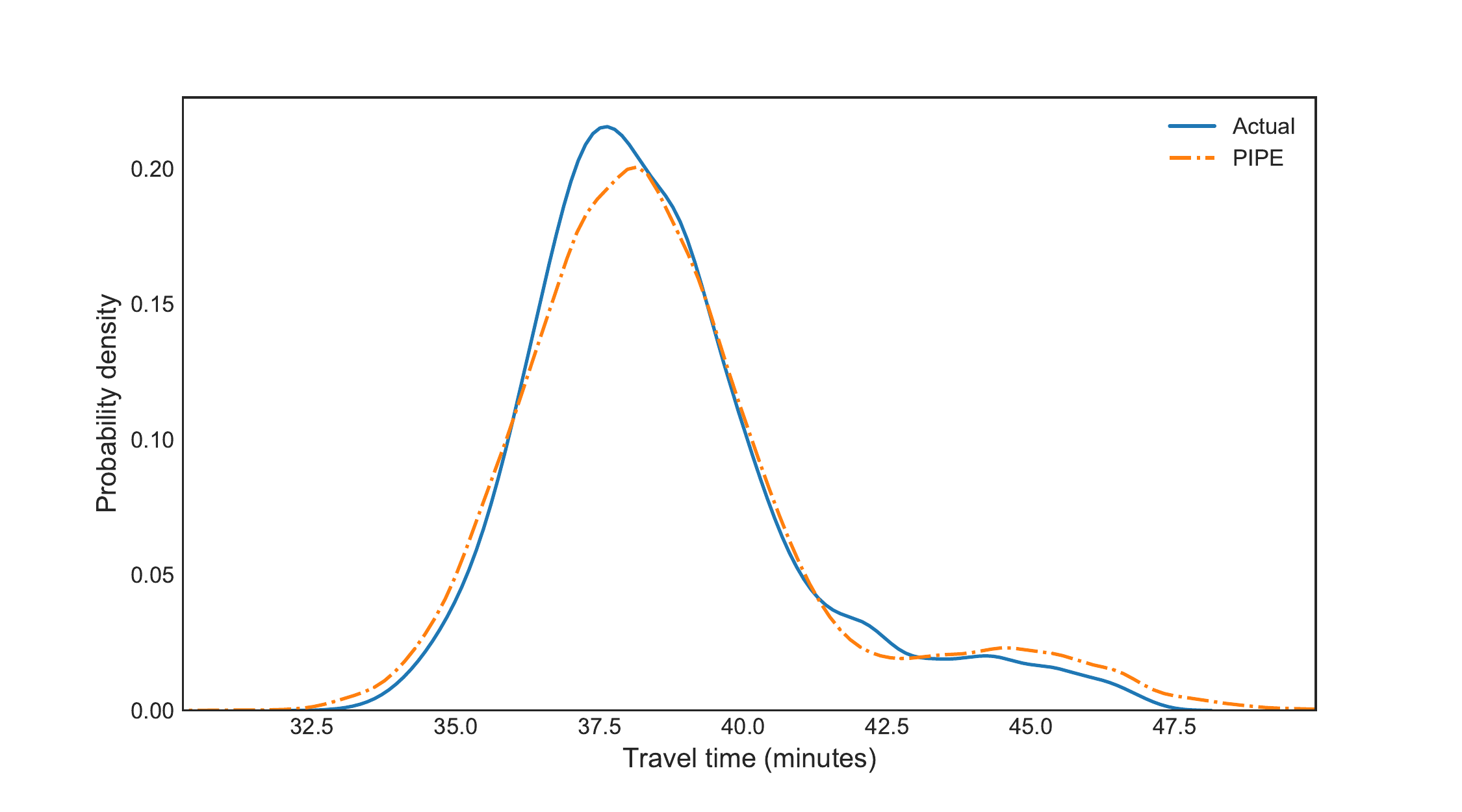}}
    \subfigure[Yew Tee to City Hall]{
        \includegraphics[width=0.3\textwidth]{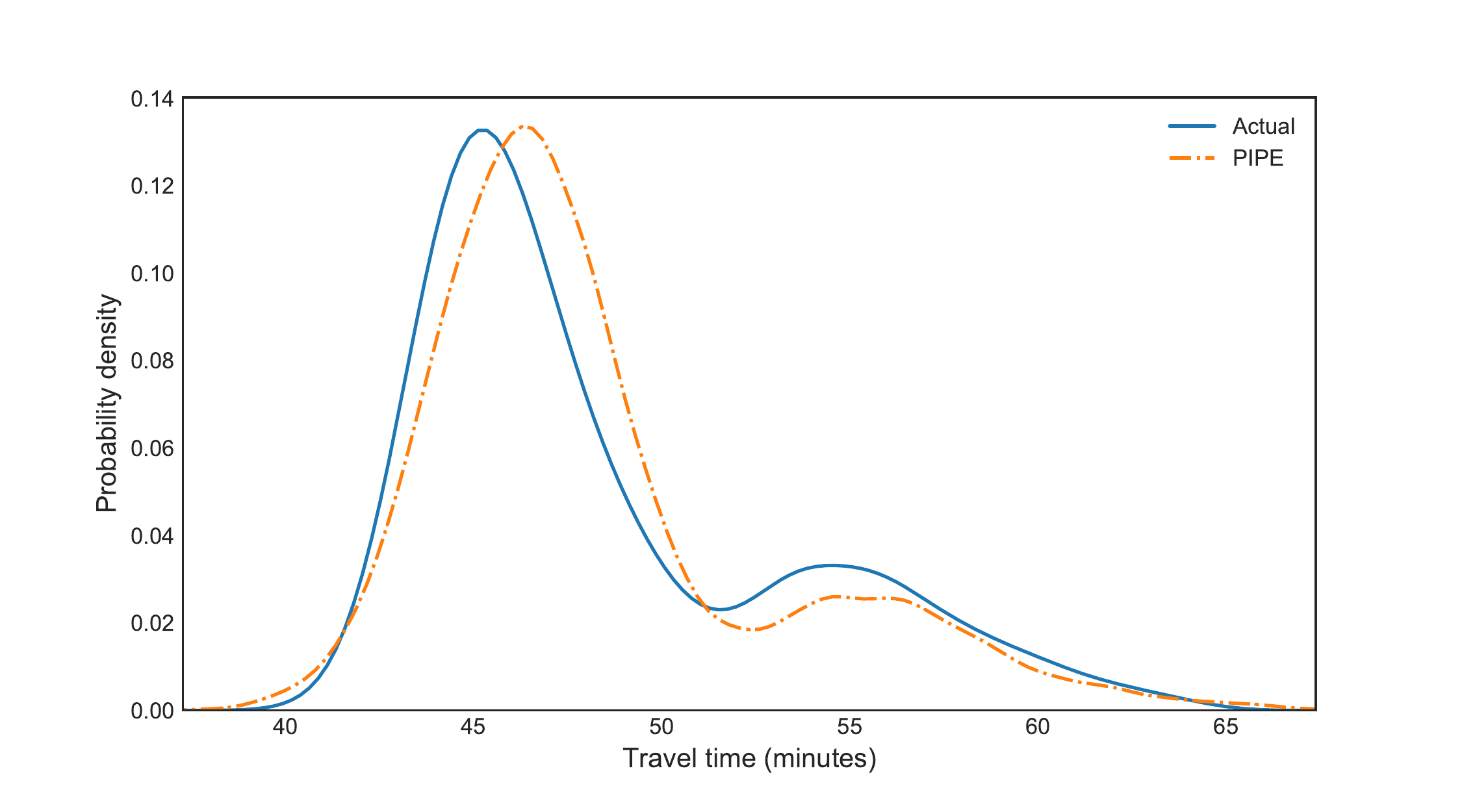} }
    \vspace{-0.1in}
    \caption{Travel time fit for multi-route OD pairs}
    \label{fig:travel_time_fit_multi}
    \vspace{-0.15in}
\end{figure*}

Similar as OD matrix prediction, here we use the 70\% training data to infer the travel time distribution of all the transit links based on the proposed truncated Gaussian mixture model. For demonstration purpose, we decompose the route from Tanah Merah station to Changi Airport station in Singapore, and report the travel time distribution for each of its transit links in Fig.~\ref{fig:travel_time_fit_link}. Note that both Tanah Merah station and Changi Airport station are located at the Changi Airport Branch (CG) line, as shown in Fig.~\ref{fig:mrt_map}. They are EW4 and CG2 respectively, with a distance of 2 segment away along the CG line. There are in total four travel links that contribute to the travel time from Tanah Merah station to Changi Airport station, as reported in Fig.~\ref{fig:travel_time_fit_link}, including a) entry link at Tanan Merah station, b) in-train travel link from Tanan Merah (EW4) to Expo station (CG1), c) in-train travel link from Expo (CG1) to Changi Airport station (CG2), d) exit link at Changi Airport station. We notice that the mean of transit link a) is relatively larger than that of other stations, this is because that CG line has a higher train headway (i.e., 6-9 minutes) than other service lines (i.e., 2-6 minutes) and consequently passengers at EW4 usually spend more time waiting for trains. We also notice that transit link c) takes much longer time than transit link b), this is because the physical distance between CG1 and CG2 is much longer than that between EW4 and CG1.

Based on the estimated travel time distribution of each transit link, we can infer the travel time distribution for each OD pair. We first present results of some OD pairs with single route (i.e., $|R_{od}|=1$) in Fig.~\ref{fig:travel_time_fit_single}, where the blue solid line is the empirical distribution of the travel time calculated by kernel density estimation based on the training data set, the orange dash line is the probability density function of our estimated truncated Gaussian distribution. 
As observed, PIPE is able to provide a nice fit of the empirical travel time distribution. 

We next present the results corresponding to OD pairs with multiple routes in Fig.~\ref{fig:travel_time_fit_multi}. Different from the results in Fig.~\ref{fig:travel_time_fit_single}, we could observe multiple modes, with each representing the travel time distribution of one particular route and the magnitude of the mode value being proportional to its route choice probability. As we can observe, PIPE is able to provide accurate predictions on both travel time and route choice probabilities. 

\begin{figure*}[!htb]
    \centering
    \subfigure[Admiralty to Farrer Park]{\label{fig:od_matrix_1}
        \includegraphics[width=0.3\textwidth]{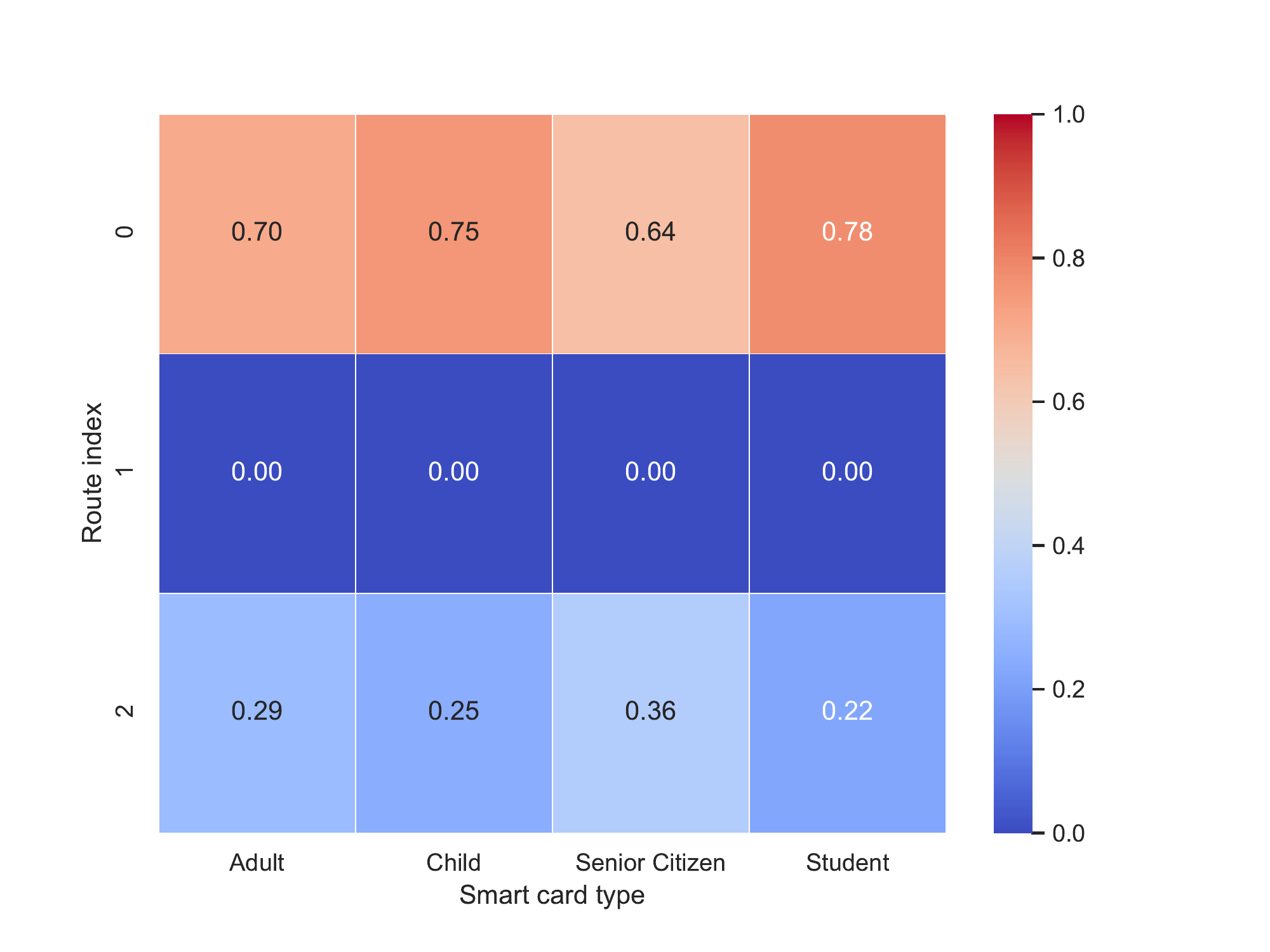}}
    \subfigure[Chinatown to Raffles Place]{
        \includegraphics[width=0.3\textwidth]{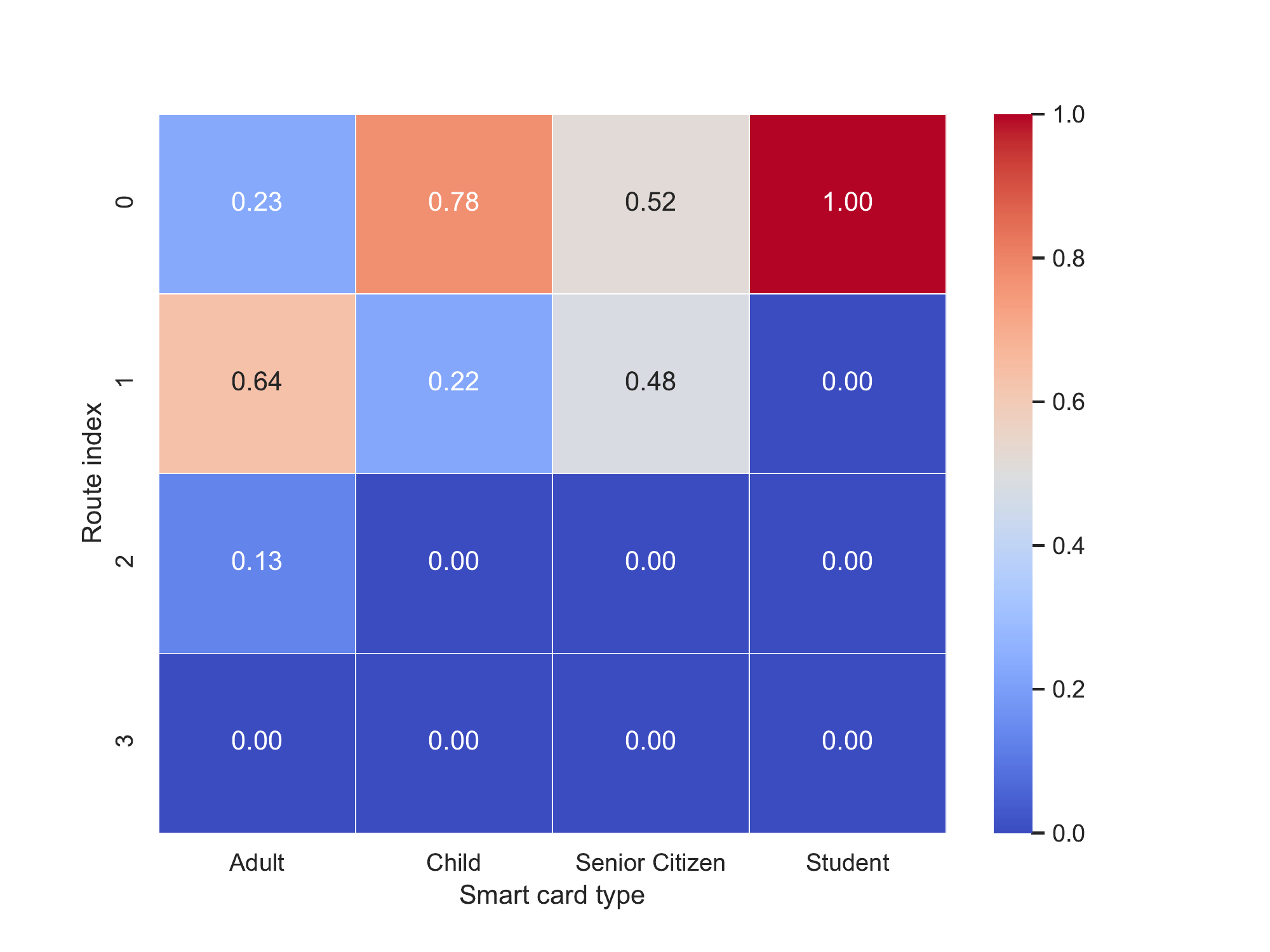}}
    \subfigure[Ang Mo Kio to Kent Ridge]{
        \includegraphics[width=0.3\textwidth]{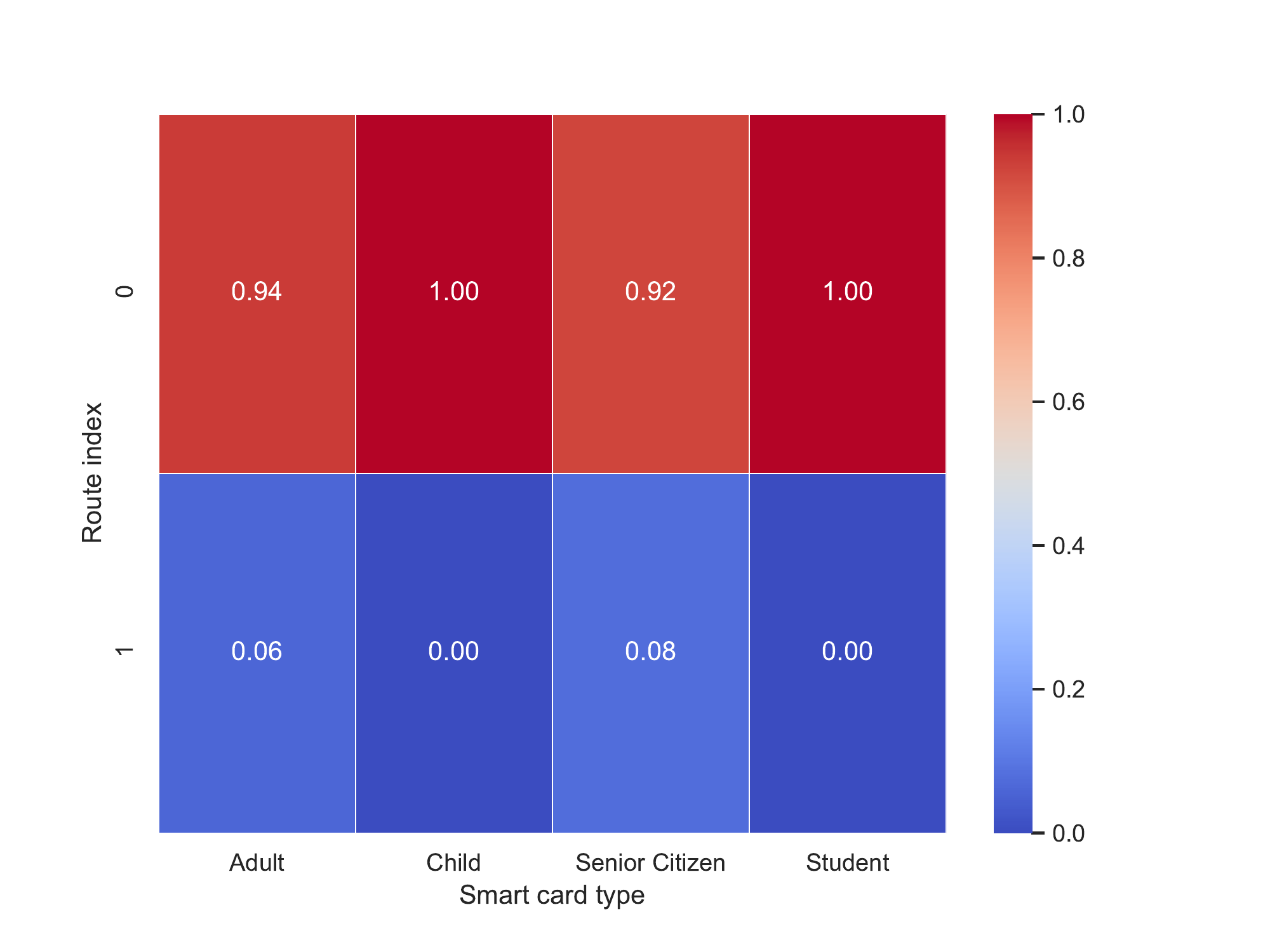}}
    \vspace{-0.1in}
    \caption{Route choice preferences by smart card type}
    \label{fig:route_preferences}
    \vspace{-0.15in}
\end{figure*}

Furthermore, we analyze route preferences of different smart card types, which are captured by the weights of truncated Gaussians in the mixture model. Fig.~\ref{fig:route_preferences} shows route choice probabilities of different smart card types for three selected OD pairs. The numerical value in each cell represents the estimated $\pi_{od}^{c}$. It shows that in Fig.~\ref{fig:route_preferences} (a), where there are three candidate routes bringing passengers from Admiralty station to Farrer Park station, different smart card types share similar route choice preferences. Most of the passengers prefer route $0$, only small ratio of passengers take route $2$, while no passengers are willing to use route $1$ to complete their trips. In contrast, in Fig.~\ref{fig:route_preferences} (b), the route choice preferences varies across smart card types. Both child and student passenger prefer route $0$, while adults like route $1$ more, and senior citizens like route $0$ and route $1$ almost equally. Another interesting phenomenon we observed is, in many cases, even there are multiple candidate routes available, some of them are rarely or never travelled by people. For example, as shown in Fig.~\ref{fig:route_preferences} (c), almost all passengers take route $0$ and very few passengers take route $1$ when they travel from Ang Mo Kio station to Kent Ridge station.

\subsubsection{Alighting Rate Forecasting}
\label{subsubsec:alighting_rate}
\begin{figure*}[!htb]
    \centering
    \subfigure[Bartley]{
        \includegraphics[width=0.3\textwidth]{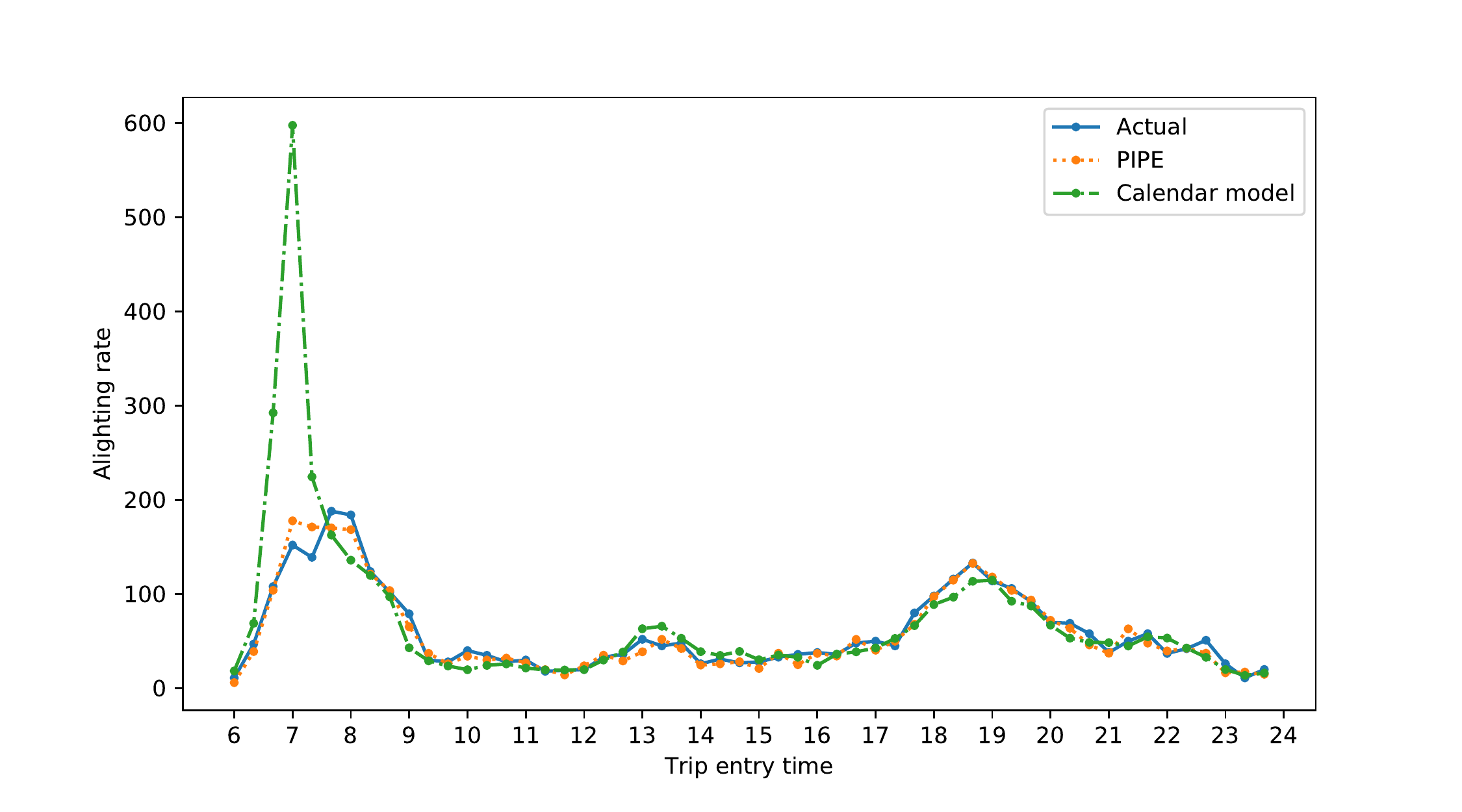}}
    \subfigure[City Hall]{
        \includegraphics[width=0.3\textwidth]{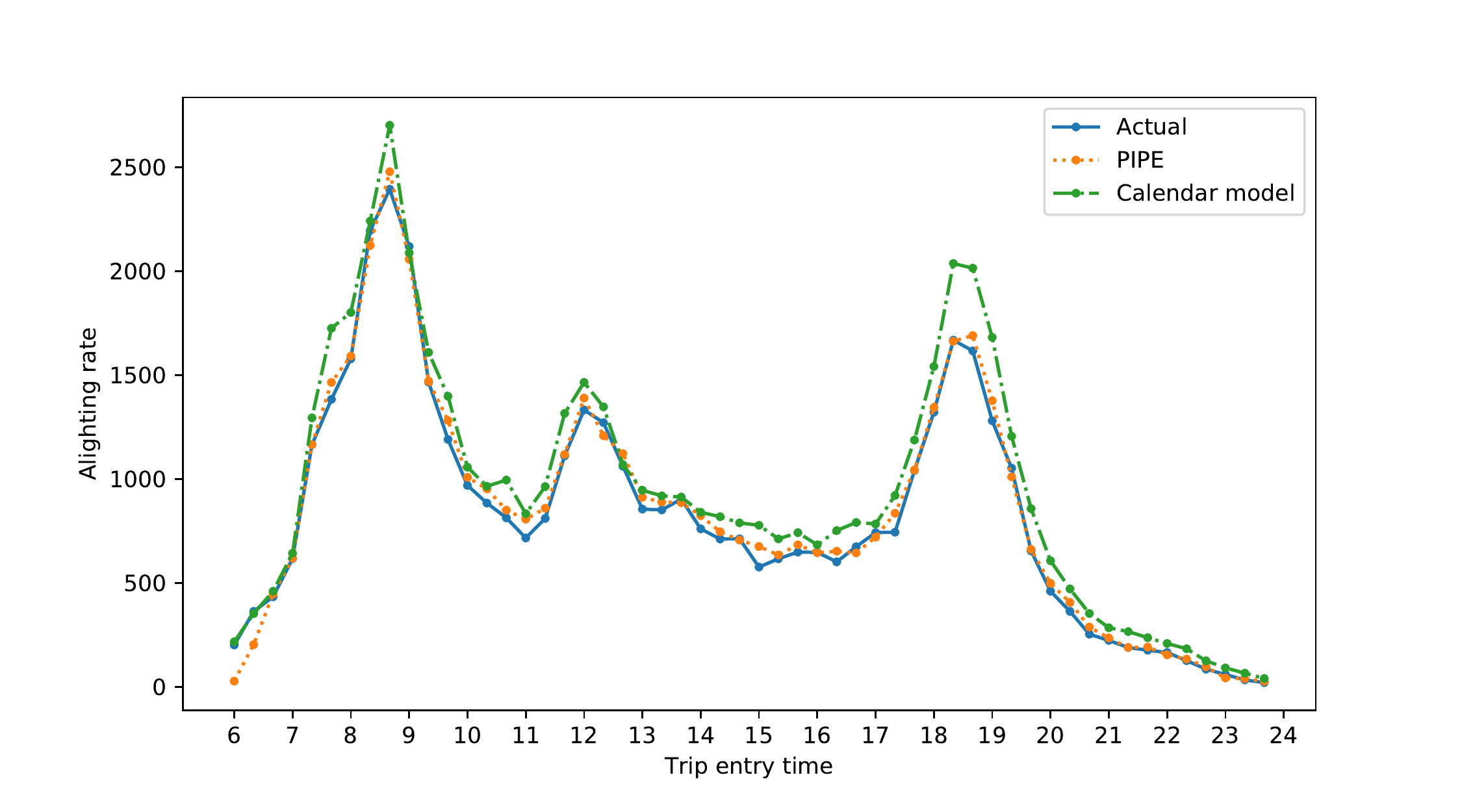}}
    \subfigure[Punggol]{
        \includegraphics[width=0.3\textwidth]{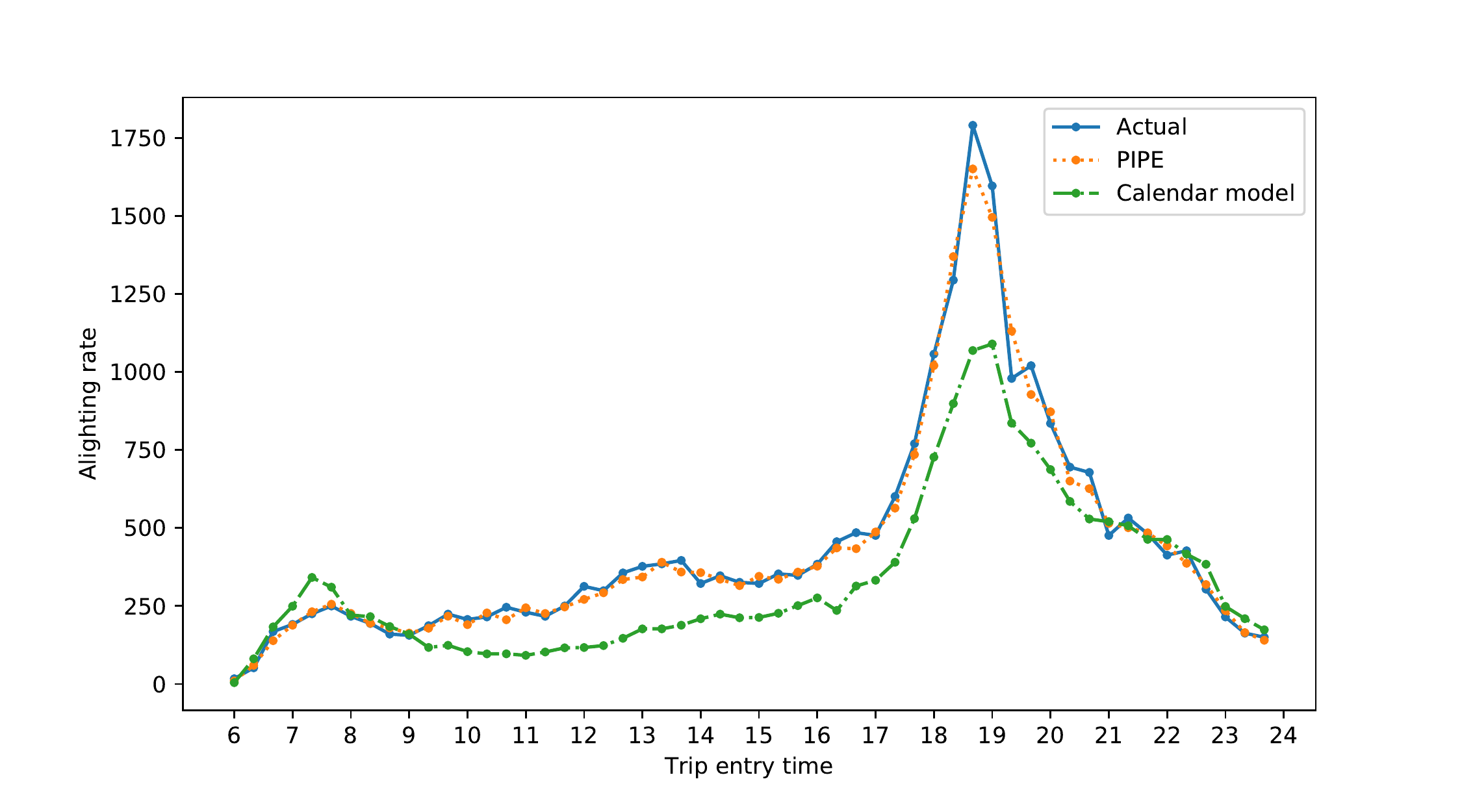}}
    \vspace{-0.1in}
    \caption{Alighting rate prediction results}
    \label{fig:alight_rate}
    \vspace{0.1in}
\end{figure*}

Now based on the estimation of OD matrix, travel time parameters and route choice probabilities, we use the remaining 10\% dataset to evaluate the prediction performance of PIPE. 
We first use PIPE to predict the alighting rate (a.k.a., outflow) at each station based on Equation~\eqref{eq:tap_out_flow}, whose ground truth is known. Take one-step ahead prediction results as an example. Fig.~\ref{fig:alight_rate} reports the predicted alighting rate of PIPE for three selected stations. The actual outflow together with the prediction result of the vanilla model are also reported. As can be observed, PIPE can track the small temporal fluctuations of outflow of the testing day well, while vanilla model fails to capture lots of details, such as the magnitudes of the peak hour. Though the performance of PIPE deteriorate as the prediction ahead step increases, it still performs consistently better than the vanilla method. Furthermore, we consider directly using random forest to predict $X_{i}^{out}(\tau)$ by taking $X_{i}^{in}(\tau-l), X_{i}^{out}(\tau-l), l=1,\ldots,\Delta$ as input. This can be regarded as another baseline of PIPE, which uses a black box model for alighting rate prediction, without considering the travel behaviors or traces of passengers inside the metro system. As shown in Table~\ref{tab:alighting_rate}, PIPE outperforms random forest by a large extent in terms of MSE, which demonstrates the efficacy and advantages of our white-box model.

\renewcommand{\tabcolsep}{1pt}
\begin{table}[t!]
\centering
\caption{Alighting Rate Prediction Error (MSE)}
\label{tab:alighting_rate}
\begin{tabular}{|p{4.3cm}|p{1cm}|p{1cm}|p{1cm}|}
\hline
\diagbox[]{Model}{Ahead time} & \textbf{1} & \textbf{4} & \textbf{6}
\\ \hline
Calendar model & 6840.64 & 6840.64 & 6840.64
\\ \hline
Random Forest & 3328.21 & 4826.65 & 5268.05
\\ \hline
PIPE & \textbf{2433.20} & \textbf{2591.15} & \textbf{2963.79}
\\ \hline
\end{tabular}
\vspace{-0.15in}
\end{table}

\subsubsection{In-situ Passenger Density Prediction}


\begin{figure*}[t!]
    \centering
    \subfigure[To east direction]{
        \includegraphics[width=0.49\textwidth]{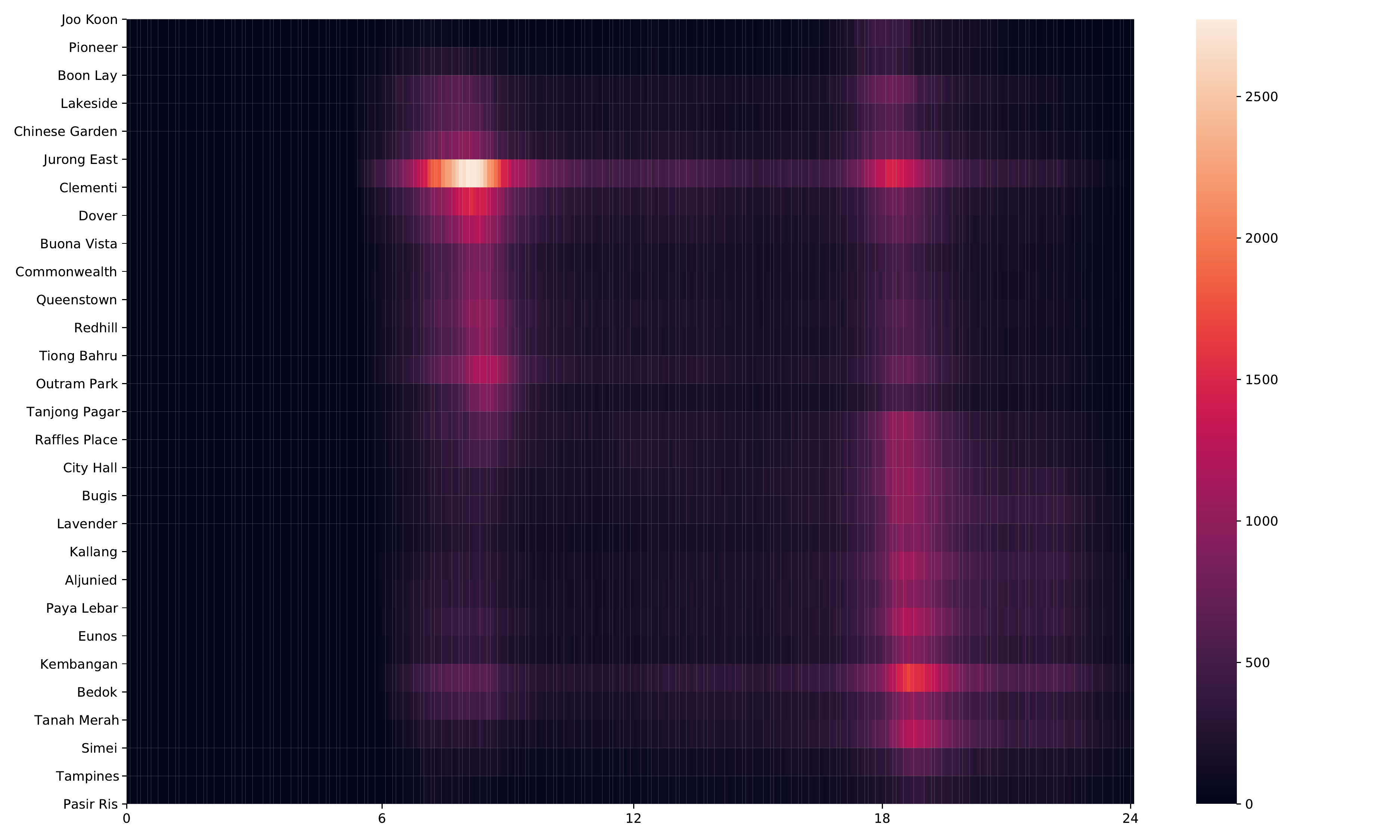}}
    \subfigure[To west direction]{
        \includegraphics[width=0.49\textwidth]{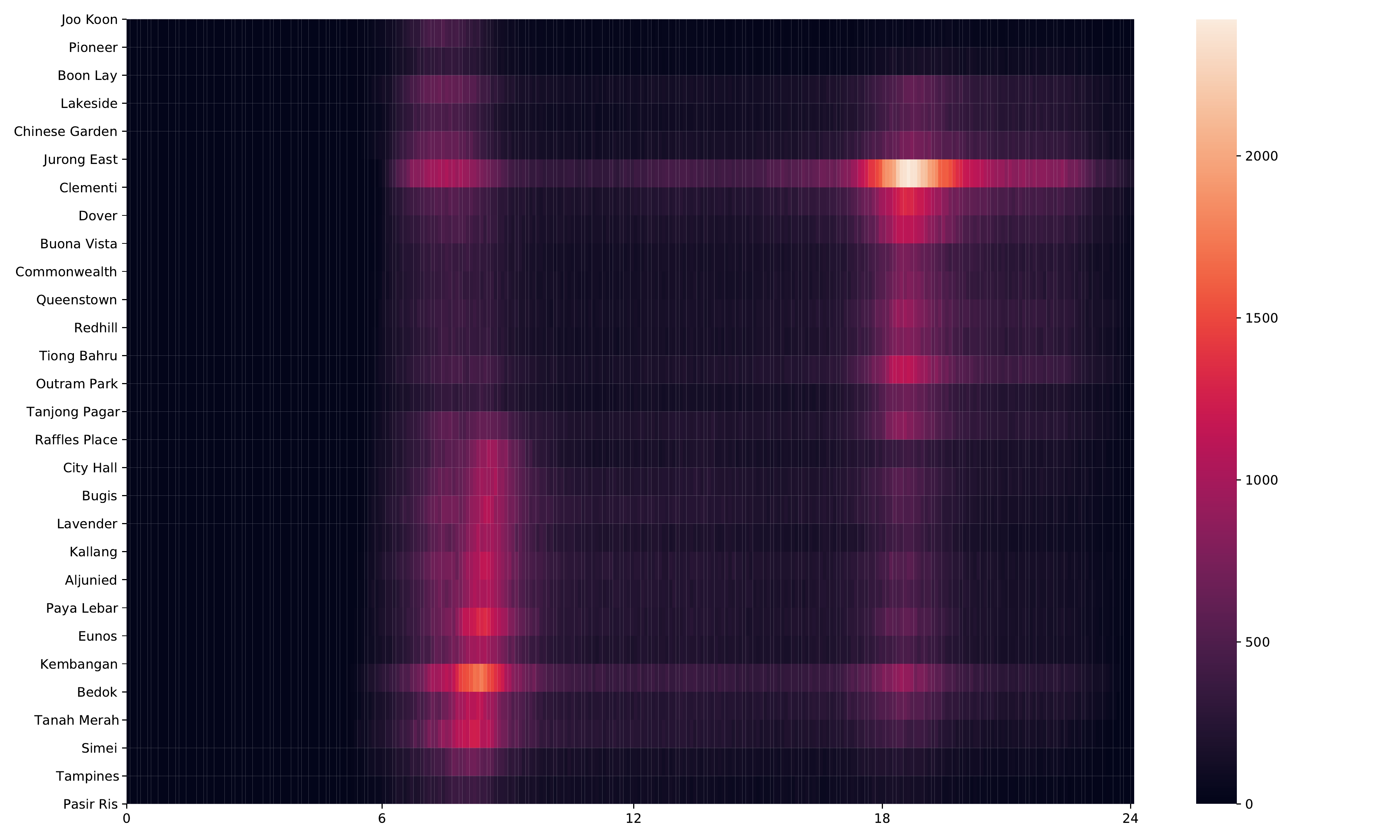}}
    \vspace{-0.1in}
    \caption{East-West line passenger density}
    \label{fig:east_west_line}
    \vspace{-0.1in}
\end{figure*}

Last but not least, we visualize the prediction results of in-situ passenger density. It is noted that we cannot achieve the ground truth data for this part. We first choose the most busy line, i.e., East-West (EW) line to illustrate the passenger density of different segments over time. For better spatio-temporal pattern illustration, we show the passenger densities along two train directions, i.e., to east direction and to west direction, separately. As shown in Fig.~\ref{fig:east_west_line}, there are two commuting peak periods, one is the morning peak hour occurring around 8am, when people left home and make trip to office. Therefore we can observe a morning peak originating from western residential districts (e.g., Jurong East station) all the way to central business districts (CBD) (e.g., City Hall station) in Fig.~\ref{fig:east_west_line} (a), and another morning peak originating from eastern residential areas (e.g., Bedok Station) all the way to CBD in Fig.~\ref{fig:east_west_line} (b).
The other peak hour occurs at about 6pm, when people 
get off work and make trip home or go to places for entertainment activities like dinner and shopping. Consequently, we can observe a evening peak originating from CBD to eastern residential districts in Fig.~\ref{fig:east_west_line} (a), and another evening peak originating from CBD to western residential areas in Fig.~\ref{fig:east_west_line} (b).

\begin{figure}[t!]
    \centering
    \includegraphics[width=0.4\textwidth]{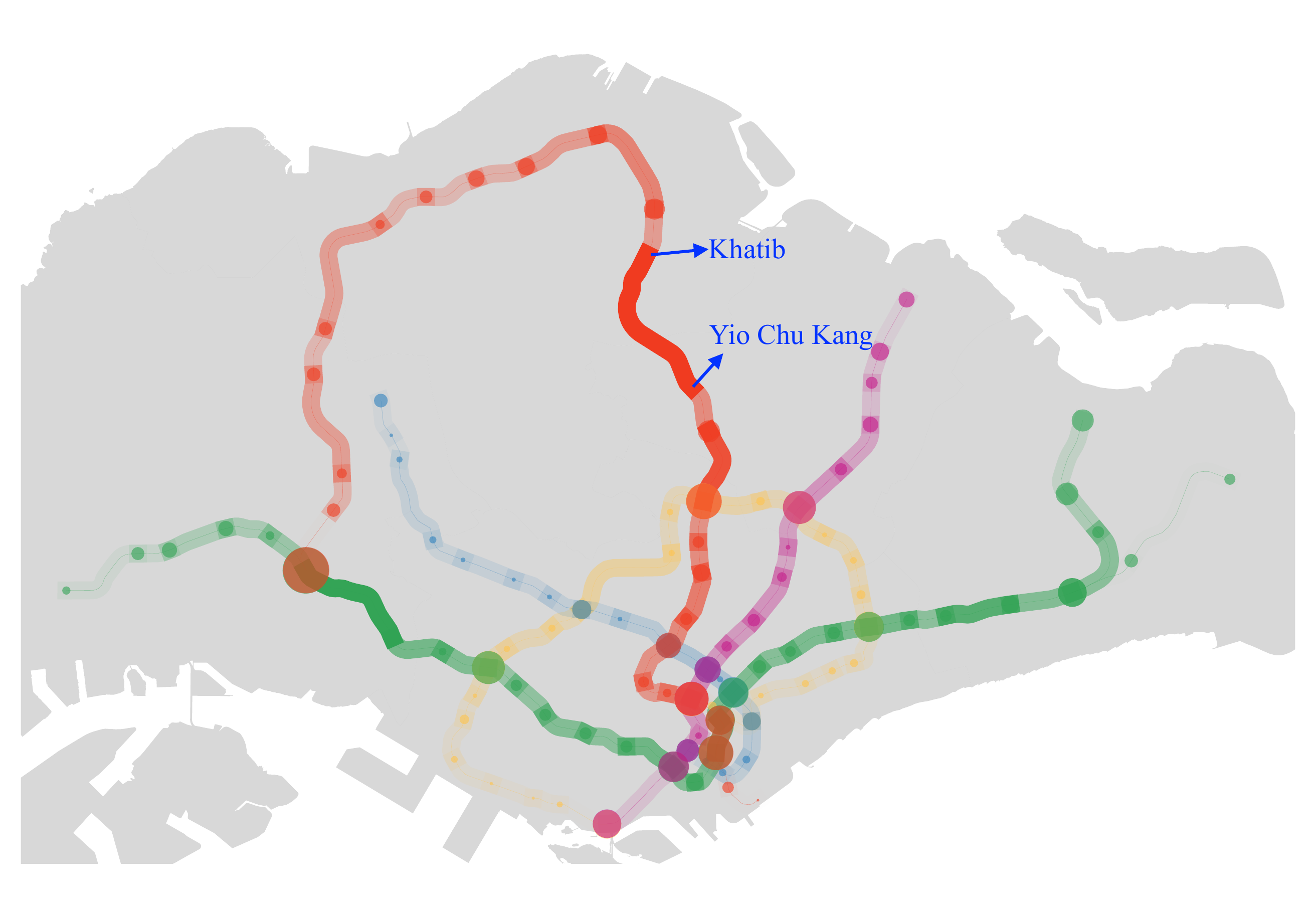}
    \caption{In-situ passenger density @ 9am}
    \label{fig:in_situ_passengers}
    \vspace{-0.15in}
\end{figure}

Then we evaluate the spatial distribution of passenger density by reporting in-situ passenger density snapshots of the whole metro system at particular time point. For example, Fig.~\ref{fig:in_situ_passengers} reports the passenger density distribution of the whole Singapore MRT system at 9am. Here the node size indicates the number of passengers inside each MRT station at the corresponding timestamp, and color intensity of each segment is proportional to the number of passengers traveling on the segment. Obviously there are a few MRT stations serving larger number of passengers than the rest, most of these busy stations are transit hubs where people make transfers or stop by for activities like dining. We also notice that the most crowded edge is the one between Yio Chu Kang station and Khatib station, this is because this edge is the longest in the MRT system, and in most of the time there are two trains running on the same edge, while for other edges, in most of the time, there is only one single train.
\section{Conclusion}
\label{sec:conclusion}
\vspace{-0.05in}

In this paper, we proposed a statistical inference framework \emph{PIPE} that makes fine-grain and accurate in-situ passenger densities prediction across the metro network. \emph{PIPE} conducted inference tasks including time-dependent OD matrices forecasting, study of travel time distribution of any travel link inside the metro network, and inference of route choice probabilities. Based on derived parameters from the inference tasks, we further estimate the passenger flow properties in terms of alighting rate at each metro station and in-situ passenger density distribution. We apply our solution in Singapore MRT network, and the satisfactory prediction performance demonstrates its applicability and efficiency.


\bibliographystyle{IEEEtran}
\bibliography{main}

\begin{thebibliography}{10}
\providecommand{\url}[1]{#1}
\csname url@samestyle\endcsname
\providecommand{\newblock}{\relax}
\providecommand{\bibinfo}[2]{#2}
\providecommand{\BIBentrySTDinterwordspacing}{\spaceskip=0pt\relax}
\providecommand{\BIBentryALTinterwordstretchfactor}{4}
\providecommand{\BIBentryALTinterwordspacing}{\spaceskip=\fontdimen2\font plus
\BIBentryALTinterwordstretchfactor\fontdimen3\font minus
  \fontdimen4\font\relax}
\providecommand{\BIBforeignlanguage}[2]{{%
\expandafter\ifx\csname l@#1\endcsname\relax
\typeout{** WARNING: IEEEtran.bst: No hyphenation pattern has been}%
\typeout{** loaded for the language `#1'. Using the pattern for}%
\typeout{** the default language instead.}%
\else
\language=\csname l@#1\endcsname
\fi
#2}}
\providecommand{\BIBdecl}{\relax}
\BIBdecl

\bibitem{pelletier2011smart}
M.-P. Pelletier, M.~Tr{\'e}panier, and C.~Morency, ``Smart card data use in
  public transit: A literature review,'' \emph{Transportation Research Part C:
  Emerging Technologies}, vol.~19, no.~4, pp. 557--568, 2011.

\bibitem{heydenrijk2018supervised}
L.~Heydenrijk-Ottens, V.~Degeler, D.~Luo, N.~van Oort, and J.~van Lint,
  ``Supervised learning: Predicting passenger load in public transport,'' in
  \emph{CASPT}, 2018.

\bibitem{vandewiele2017predicting}
G.~Vandewiele, P.~Colpaert, O.~Janssens, J.~Van~Herwegen, R.~Verborgh,
  E.~Mannens, F.~Ongenae, and F.~De~Turck, ``Predicting train occupancies based
  on query logs and external data sources,'' in \emph{WWW'17 Companion}, pp.
  1469--1474.

\bibitem{pasini2019lstm}
K.~Pasini, M.~Khouadjia, A.~Same, F.~Ganansia, and L.~Oukhellou, ``Lstm
  encoder-predictor for short-term train load forecasting,'' in \emph{ECML PKDD
  2019}, pp. 535--551.

\bibitem{Hong2017}
L.~Hong, W.~Li, and W.~Zhu, ``Assigning passenger flows on a metro network
  based on automatic fare collection data and timetable,'' \emph{Discrete
  Dynamics in Nature and Society}.

\bibitem{xu2018railway}
X.~Xu, Y.~Dou, Z.~Zhou, T.~Liao, Y.~Lu, and Y.~Tan, ``Railway passenger flow
  forecasting based on time series analysis with big data,'' in
  \emph{CCDC}.\hskip 1em plus 0.5em minus 0.4em\relax IEEE, 2018, pp.
  3584--3590.

\bibitem{chen2019subway}
E.~Chen, Z.~Ye, C.~Wang, and M.~Xu, ``Subway passenger flow prediction for
  special events using smart card data,'' \emph{IEEE Transactions on ITS},
  vol.~21, no.~3, pp. 1109--1120, 2019.

\bibitem{gong2018network}
Y.~Gong, Z.~Li, J.~Zhang, W.~Liu, Y.~Zheng, and C.~Kirsch, ``Network-wide crowd
  flow prediction of sydney trains via customized online non-negative matrix
  factorization,'' in \emph{ICKM}, 2018, pp. 1243--1252.

\bibitem{sun2015novel}
Y.~Sun, B.~Leng, and W.~Guan, ``A novel wavelet-svm short-time passenger flow
  prediction in beijing subway system,'' \emph{Neurocomputing}, vol. 166, pp.
  109--121, 2015.

\bibitem{li2017forecasting}
Y.~Li, X.~Wang, S.~Sun, X.~Ma, and G.~Lu, ``Forecasting short-term subway
  passenger flow under special events scenarios using multiscale radial basis
  function networks,'' \emph{Transportation Research Part C: Emerging
  Technologies}, vol.~77, pp. 306--328, 2017.

\bibitem{lv2014traffic}
Y.~Lv, Y.~Duan, W.~Kang, Z.~Li, and F.-Y. Wang, ``Traffic flow prediction with
  big data: a deep learning approach,'' \emph{IEEE Transactions on ITS},
  vol.~16, no.~2, pp. 865--873, 2014.

\bibitem{liu2019deeppf}
Y.~Liu, Z.~Liu, and R.~Jia, ``Deeppf: A deep learning based architecture for
  metro passenger flow prediction,'' \emph{Transportation Research Part C:
  Emerging Technologies}, vol. 101, pp. 18--34, 2019.

\bibitem{hao2019sequence}
S.~Hao, D.-H. Lee, and D.~Zhao, ``Sequence to sequence learning with attention
  mechanism for short-term passenger flow prediction in large-scale metro
  system,'' \emph{Transportation Research Part C: Emerging Technologies}, vol.
  107, pp. 287--300, 2019.

\bibitem{ashok2002estimation}
K.~Ashok and M.~E. Ben-Akiva, ``Estimation and prediction of time-dependent
  origin-destination flows with a stochastic mapping to path flows and link
  flows,'' \emph{Transportation Science}, vol.~36, no.~2, pp. 184--198, 2002.

\bibitem{chen2011short}
X.~Chen, S.~Guo, L.~Yu, and B.~Hellinga, ``Short-term forecasting of transit
  route od matrix with smart card data,'' in \emph{ITSC}.\hskip 1em plus 0.5em
  minus 0.4em\relax IEEE, 2011, pp. 1513--1518.

\bibitem{van2012dynamic}
E.~Van~der Hurk, L.~G. Kroon, G.~Mar{\'o}ti, and P.~Vervest, ``Dynamic forecast
  model of time dependent passenger flows for disruption management,'' in
  \emph{CASPT}, 2012, pp. 23--27.

\bibitem{toque2016forecasting}
F.~Toqu{\'e}, E.~C{\^o}me, M.~K. El~Mahrsi, and L.~Oukhellou, ``Forecasting
  dynamic public transport origin-destination matrices with long-short term
  memory recurrent neural networks,'' in \emph{ITSC}.\hskip 1em plus 0.5em
  minus 0.4em\relax IEEE, 2016, pp. 1071--1076.

\bibitem{hu2020stochastic}
J.~Hu, B.~Yang, C.~Guo, C.~S. Jensen, and H.~Xiong, ``Stochastic
  origin-destination matrix forecasting using dual-stage graph convolutional,
  recurrent neural networks,'' in \emph{ICDE}.\hskip 1em plus 0.5em minus
  0.4em\relax IEEE, 2020, pp. 1417--1428.

\bibitem{shi2020predicting}
H.~Shi, Q.~Yao, Q.~Guo, Y.~Li, L.~Zhang, J.~Ye, Y.~Li, and Y.~Liu, ``Predicting
  origin-destination flow via multi-perspective graph convolutional network,''
  in \emph{ICDE}.\hskip 1em plus 0.5em minus 0.4em\relax IEEE, 2020, pp.
  1818--1821.

\bibitem{zhao2016estimation}
J.~Zhao, F.~Zhang, L.~Tu, C.~Xu, D.~Shen, C.~Tian, X.-Y. Li, and Z.~Li,
  ``Estimation of passenger route choice pattern using smart card data for
  complex metro systems,'' \emph{IEEE Transactions on ITS}, vol.~18, no.~4, pp.
  790--801, 2016.

\bibitem{Sun2015}
L.~Sun, Y.~Lu, J.~G. Jin, D.-H. Lee, and K.~W. Axhausen, ``An integrated
  bayesian approach for passenger flow assignment in metro networks,''
  \emph{Transportation Research Part C: Emerging Technologies}, vol.~52, pp.
  116 -- 131, 2015.

\bibitem{xu2018learning}
X.~Xu, L.~Xie, H.~Li, and L.~Qin, ``Learning the route choice behavior of
  subway passengers from afc data,'' \emph{Expert Systems with Applications},
  vol.~95, pp. 324--332, 2018.

\bibitem{tian2020tripdecoder}
X.~Tian, B.~Zheng, Y.~Wang, H.-T. Huang, and C.-C. Hung, ``Tripdecoder: Study
  travel time attributes and route preferences of metro systems from smart card
  data,'' \emph{arXiv preprint arXiv:2005.01492}, 2020.

\bibitem{NIPS2017}
N.~Colombo, R.~Silva, and S.~M. Kang, ``Tomography of the london underground: a
  scalable model for origin-destination data,'' in \emph{NIPS}, 2017, pp.
  3062--3073.

\bibitem{sun2012using}
L.~Sun, D.-H. Lee, A.~Erath, and X.~Huang, ``Using smart card data to extract
  passenger's spatio-temporal density and train's trajectory of mrt system,''
  in \emph{SIGKDD international workshop on urban computing}, 2012, pp.
  142--148.

\bibitem{zhang2015spatiotemporal}
F.~Zhang, J.~Zhao, C.~Tian, C.~Xu, X.~Liu, and L.~Rao, ``Spatiotemporal
  segmentation of metro trips using smart card data,'' \emph{IEEE Transactions
  on Vehicular Technology}, vol.~65, no.~3, pp. 1137--1149, 2015.

\bibitem{jenelius2019data}
E.~Jenelius, ``Data-driven metro train crowding prediction based on real-time
  load data,'' \emph{IEEE Transactions on ITS}, vol.~21, no.~6, pp. 2254--2265,
  2019.

\bibitem{tibshirani1996regression}
R.~Tibshirani, ``Regression shrinkage and selection via the lasso,''
  \emph{Journal of the Royal Statistical Society: Series B (Methodological)},
  vol.~58, no.~1, pp. 267--288, 1996.

\bibitem{hoerl1970ridge}
A.~E. Hoerl and R.~W. Kennard, ``Ridge regression: Biased estimation for
  nonorthogonal problems,'' \emph{Technometrics}, vol.~12, no.~1, pp. 55--67,
  1970.

\bibitem{liaw2002classification}
A.~Liaw, M.~Wiener \emph{et~al.}, ``Classification and regression by
  randomforest,'' \emph{R news}, vol.~2, no.~3, pp. 18--22, 2002.

\bibitem{makridakis1997arma}
S.~Makridakis and M.~Hibon, ``Arma models and the box--jenkins methodology,''
  \emph{Journal of Forecasting}, vol.~16, no.~3, pp. 147--163, 1997.

\bibitem{hochreiter1997long}
S.~Hochreiter and J.~Schmidhuber, ``Long short-term memory,'' \emph{Neural
  computation}, vol.~9, no.~8, pp. 1735--1780, 1997.

\bibitem{prechelt1998early}
L.~Prechelt, ``Early stopping-but when?'' in \emph{Neural Networks: Tricks of
  the trade}.\hskip 1em plus 0.5em minus 0.4em\relax Springer, 1998, pp.
  55--69.

\bibitem{wang2012speed}
Y.~Wang, W.~Dong, L.~Zhang, D.~Chin, M.~Papageorgiou, G.~Rose, and W.~Young,
  ``Speed modeling and travel time estimation based on truncated normal and
  lognormal distributions,'' \emph{Transportation research record}, vol. 2315,
  no.~1, pp. 66--72, 2012.

\bibitem{Cozman-1994-13767}
F.~Cozman and E.~Krotkov, ``Truncated gaussians as tolerance sets,'' Carnegie
  Mellon University, Pittsburgh, PA, Tech. Rep. CMU-RI-TR-94-35, September
  1994.

\end{thebibliography}

\end{document}